\documentclass{article}

\usepackage{arxiv}

\usepackage[utf8]{inputenc} 
\usepackage[T1]{fontenc}    
\usepackage{hyperref}       
\usepackage{url}            
\usepackage{booktabs}       
\usepackage{array}
\usepackage{amsfonts}       
\usepackage{nicefrac}       
\usepackage{graphicx}
\usepackage{subcaption}
\usepackage{multirow}
\usepackage{adjustbox}
\usepackage[numbers,sort&compress]{natbib}
\usepackage{doi}
\usepackage{amsmath}
\usepackage{amssymb}
\usepackage{algorithm}
\usepackage{algpseudocode}
\usepackage{xcolor}
\usepackage{fancyhdr}
\usepackage{amsthm}
\usepackage{cleveref}
\usepackage{bm}
\usepackage[mathscr]{euscript}
\usepackage{textcomp}

\DeclareMathAlphabet{\pazocal}{OMS}{zplm}{m}{n}

\usepackage{lscape}		
\usepackage{fancyvrb}

\newtheoremstyle{remarkstyle}  
  {5pt}                        
  {5pt}                        
  {}                           
  {}                           
  {\bfseries}                  
  {.}                          
  { }                          
  {}                           
\newtheorem{theorem}{Theorem}[section]

\theoremstyle{remarkstyle}
\newtheorem{remark}[theorem]{Remark}

\crefname{figure}{Fig.}{Figs.}
\Crefname{figure}{Figure}{Figures}
\crefname{algorithm}{Algorithm}{Algorithms}
\crefname{equation}{Eq.}{Eqs.}

\title{
A Multimodal Conditional Mixture Model with Distribution-Level Physics Priors
}

\author{
Jinkyo Han \\
Department of Mechanical Engineering \\
Northwestern University \\
\texttt{jinkyohan2030@u.northwestern.edu}
\And
Bahador Bahmani\thanks{Corresponding author.}\\
Department of Mechanical Engineering \\
Northwestern University \\
\texttt{bahador.bahmani@northwestern.edu}
}

\renewcommand{\shorttitle}{Multimodal Mixture Model with Physics Prior}

\hypersetup{
    colorlinks=true,      
    linkcolor=green,      
    urlcolor=blue,        
    citecolor=blue        
}

\begin{document}
\maketitle

\begin{abstract}
Many scientific and engineering systems exhibit intrinsically multimodal behavior arising from latent regime switching and non-unique physical mechanisms. In such settings, learning the full conditional distribution of admissible outcomes in a physically consistent and interpretable manner remains a challenge. While recent advances in machine learning have enabled powerful multimodal generative modeling, their integration with physics-constrained scientific modeling remains nontrivial, particularly when physical structure must be preserved or data are limited.
This work develops a physics-informed multimodal conditional modeling framework based on mixture density representations. Mixture density networks (MDNs) provide an explicit and interpretable parameterization of multimodal conditional distributions. Physical knowledge is embedded through component-specific regularization terms that penalize violations of governing equations or physical laws. This formulation naturally accommodates non-uniqueness and stochasticity while remaining computationally efficient and amenable to conditioning on contextual inputs.
The proposed framework is evaluated across a range of scientific problems in which multimodality arises from intrinsic physical mechanisms rather than observational noise, including bifurcation phenomena in nonlinear dynamical systems, stochastic partial differential equations, and atomistic-scale shock dynamics. In addition, the proposed method is compared with a conditional flow matching (CFM) model, a representative state-of-the-art generative modeling approach, demonstrating that MDNs can achieve competitive performance while offering a simpler and more interpretable formulation.
\end{abstract}

\keywords{
Mixture Density Network, Mixture of Experts, Physics Informed Neural Network, Multimodal Data
}

\section{Introduction}
Many scientific and engineering systems exhibit intrinsically multimodal behavior, arising from heterogeneous material responses, incomplete observations, latent regime switching, or non-unique mappings between inputs and outputs. Examples include bifurcation phenomena \cite{thompson1973general}, stochastic partial differential equations (SPDEs) \cite{xu2011stochastic,crabtree2024micro}, shock-driven atomistic dynamics \cite{li2017shock,branicio2018plane}. In these settings, predicting the full conditional distribution of admissible outcomes is required and, ideally, the resulting models would satisfy physical consistency and admit meaningful interpretation.

Traditional modeling approaches in computational science often rely on unimodal uncertainty representations \citep{sundararajan2012probabilistic}. Probabilistic methods \cite{liu1986probabilistic,spanos1989stochastic,papadrakakis1996structural,beck1998updating,kennedy2001bayesian,simpson2001kriging,ghanem2003stochastic,beck2004model,kaymaz2005application,marzouk2007stochastic,sudret2008global,martin2012stochastic,bilionis2012multi,farrell2015bayesian}, including Monte Carlo sampling, polynomial chaos expansions, Gaussian processes, and Bayesian inference, offer rigorous tools for uncertainty quantification and have seen broad success in practice. These methods are commonly developed under unimodal formulations with explicitly defined stochastic inputs, which motivates the exploration of complementary approaches when multimodality emerges from implicit or unresolved physical mechanisms.

Recent advances in machine learning have significantly expanded the landscape of multimodal generative modeling. Generative adversarial networks (GANs) \citep{goodfellow2014generative}, Variational autoencoders\cite{kingma2013auto}, normalizing flows\cite{papamakarios2021normalizing}, diffusion models\cite{sohl2015deep}, score-based generative methods \cite{song2020score}, and flow matching models \cite{lipman2022flow} have demonstrated remarkable success in computer vision and language modeling. 
However, their application to physics-constrained scientific problems remains challenging \cite{rixner2021probabilistic,shu2023physics,bastek2024physics,jacobsen2025cocogen}, as both the incorporation of physical structure and the development of conditional formulations are nontrivial and remain active areas of research \citep{bond2021deep}. While these models are powerful and expressive, they often require substantial data and careful training procedures, and constraining them in a manner that preserves physical structure—such as conservation laws, symmetries, or governing equations—can be nontrivial. Moreover, their latent representations are not always readily interpretable or easily manipulated within established scientific workflows.

In this work, mixture density representations \cite{mclachlan1988mixture,jacobs1991adaptive,bishop1994mixture} are revisited as a principled and underexplored alternative for physics-constrained multimodal modeling. MDNs \cite{bishop1994mixture}, originally introduced as conditional probabilistic models, approximate complex conditional distributions as finite mixtures of simple parametric components. Despite their relative simplicity, mixture models possess several features that make them particularly attractive for scientific machine learning: \textit{explicit representation of multimodality}, \textit{interpretable components}, \textit{tractable likelihoods}, and \textit{straightforward conditioning on inputs}. While mixture-based formulations have a long history, they have recently seen renewed adoption at scale in modern large language models through mixture-of-experts architectures \cite{shazeer2017outrageously,fedus2022switch}.

A physics-informed multimodal conditional modeling framework based on mixture density representations is developed to provide a natural interface between data-driven learning and physics-based modeling. Physical priors and constraints are incorporated directly into the mixture formulation through regularization terms that penalize violations of governing equations or physical laws. These constraints are imposed at the level of the conditional distribution, rather than on a single predicted trajectory or field realization. 

The framework is evaluated across a range of scientific problems in which multimodality arises from intrinsic physical mechanisms rather than observational noise, including parametric partial differential equations with non-unique solutions, stochastic partial differential equations, atomistic-scale shock dynamics, and bifurcation phenomena in nonlinear dynamical systems. Across these settings, the mixture-based formulation captures distinct solution branches, regime transitions, and structured uncertainty that are not accessible to unimodal models.

\textbf{Related work.}
MDNs have been explored in a range of engineering and applied science settings as probabilistic surrogate models. Among these, the closest to the present work is \cite{chen2022physics}, where MDNs are augmented with regularization terms motivated by physical considerations. These regularizers are primarily statistical in nature, such as constraining variance across mixture components, rather than enforcing governing equations. Moreover, the regularization terms are applied uniformly across components, without accounting for their relative contribution to the conditional distribution, and the data considered are predominantly unimodal despite the expressive capacity of MDNs. MDNs have been used as stochastic surrogate models for uncertainty quantification \cite{peng2025efficient}, including recurrent probabilistic prediction for reliability analysis of bridge expansion joints under thermal loading \cite{wang2025probabilistic} and seismic response modeling \cite{peng2026accelerating}. In design and inverse problems, mixture-based models have been applied to microstructure design \cite{yang2021general} and to inverse design of photonic structures using convolutional MDNs \cite{unni2020deep}. Ensemble formulations of MDNs have further been investigated for probabilistic wind forecasting \cite{men2016short,zhang2020improved}. There exist other approaches for probabilistic modeling with neural networks, among which Bayesian neural networks constitute one of the most established paradigms \cite{neal2012bayesian,blundell2015weight}. Such probabilistic formulations have also been combined with physics-based regularization in physics-informed settings, where physical knowledge is incorporated through additional loss terms or priors \cite{raissi2019physics,yang2021b}.

In summary, the main contributions of this work are:
\begin{enumerate}
    \item A \emph{physics-informed} modeling framework is proposed based on mixture density representations for conditional modeling, in which physics priors are imposed through a \emph{component-specific physics regularization} of the mean functions.
    \item A \emph{class-conditioning mechanism} is introduced that incorporates discrete context variables while preserving \emph{shared structure and statistical coupling} across mixture components, avoiding the loss of cross-component correlation that can arise from training separate class-specific models.
    \item A comparison with \emph{conditional flow matching} as a representative state-of-the-art generative modeling approach.
    \item The versatility of the framework is demonstrated across diverse scientific domains in which multimodality arises from intrinsic physical mechanisms rather than observational noise.
\end{enumerate}

The remainder of this work is organized as follows.
In Section~\ref{sec:formulation}, we develop the core formulation of mixture density networks for modeling conditional multimodal distributions and extend it to incorporate class-conditional information and physics-based regularization.
Section~\ref{sec:examples} presents a collection of numerical examples that demonstrate the proposed framework on physical systems exhibiting intrinsic multimodal behavior.
Finally, Section~\ref{sec:conclusion} concludes the paper and summarizes the main outcomes and limitations.

\section{Formulation}\label{sec:formulation}
Let $(\Omega,\mathcal{F},\mathbb{P})$ be a probability space.
We consider a \emph{conditional random field}
$U:\Omega \times \mathcal{X} \to \mathbb{R}$, where
$\mathcal{X}$ denotes an index set (e.g., space--time coordinates).
The collection $\{U(x)\}_{x\in\mathcal{X}}$ defines a random field on $\mathcal{X}$. We assume access to a dataset of $N$ samples $\mathcal{D} = \{(x_i,u_i)\}_{i=1}^N$
where $x_i \in \mathcal{X}$ is the evaluation index and $u_i \in \mathbb{R}$ is the observed field value. Our objective is to learn the conditional density of field evaluations $p(u|x)$, which may exhibit intrinsic multimodality. To this end, we adopt a mixture-based probabilistic representation that is both expressive and analytically tractable.

Finite mixture density models provide a general framework for representing
multimodal probability distributions by expressing a conditional density as a
convex combination of component densities.
In the conditional setting considered here, the distribution $u$ given inputs $x$ is modeled as
\begin{equation}
    p(u|x)
    =
    \sum_{m=1}^M
    \pi_m(x)\,
    \phi_m(u|x),
\end{equation}
where $\pi_m(x)\ge 0$ such that $\sum_{m=1}^M \pi_m(x)=1$, and
$\phi_m(u|x)$ denote component conditional densities.

We focus on the subclass of conditional mixture models in which
the mixture weights and the parameters of the component densities are
parameterized by a neural network.
This yields a MDN, in which all mixture parameters
are learned jointly from data.
Denoting the network parameters collectively by $\theta$, the resulting
conditional density can be written as
\begin{equation}
    p(u|x;\theta)
    =
    \sum_{m=1}^M
    \pi_m(x;\theta)\,
    \phi_m(u|x;\theta).
\end{equation}
This parameterization induces a continuous conditional random field with
respect to the input variables, which is particularly advantageous for
physics-based applications involving gradient-based constraints.

While the component conditional parametrized densities $\phi_m(u|x;\theta)$ may in principle be chosen from a broad class of distributions, throughout this work we restrict attention to parametrized Gaussian components. This choice provides analytical tractability.

Accordingly, each component density is taken to be Gaussian of the form
\begin{equation}
    \phi_m(u|x;\theta)
    =
    \frac{1}{\sigma_m(x;\theta)\sqrt{2\pi}}
    \exp\!\left(
        -\frac{\|u-\mu_m(x;\theta)\|^2}{2\,\sigma_m(x;\theta)^2}
    \right),
\end{equation}
where $\mu_m(x;\theta)$ and $\sigma_m(x;\theta)>0$ denote the component mean and standard deviation, respectively. An advantage of the mixture model with Gaussian conditional components is that the first and second conditional statistical moments of the random field admit closed-form expressions, which is particularly useful for uncertainty quantification.
\begin{align}
&\mathbb{E}[u|x]
= \sum_{m=1}^{M} \pi_m(x)\,\mu_m(x),\\
&\mathbb{E}[u^2|x]
= \sum_{m=1}^{M} \pi_m(x)\,\bigl(\sigma_m(x)^2 + \mu_m(x)^2\bigr).
\end{align}
\remark{For scalar-valued fields, the above choice corresponds to an isotropic Gaussian; extensions to multivariate outputs with structured covariance are straightforward.}

Model parameters are learned by maximum likelihood estimation. At the population level, this corresponds to minimizing the expected negative log-likelihood
\begin{equation}\label{eq:loss_nll}
    \mathcal{L}_{\mathrm{NLL}}
    =
    \mathbb{E}_{(x,u)}
    [
        -\log (
            \sum_{m=1}^M
            \pi_m(x)\,
            \phi_m(u|x)
        )
    ].
\end{equation}
which corresponds to marginalizing over the latent mixture component assignments.

In some applications, additional categorical information is available that associates each observation with a known physical regime or class. Incorporating such information as an additional source of supervision can be particularly valuable in data-scarce settings and can help mitigate common identifiability issues in mixture density models by guiding the association between mixture components and physical regimes. Let $c \in \{1,\dots,C\}$ denote a class label associated with a data point $(x,u)$, and let $g:\{1,\dots,C\}\rightarrow\{1,\dots,M\}$ be a prescribed mapping from classes to mixture components.
Rather than treating class labels as hard constraints, they are used here to provide weak guidance to the mixture assignment during training.

Specifically, the likelihood contribution of each data point is restricted to the mixture component associated with its class label, yielding the class-informed negative log-likelihood
\begin{equation}\label{eq:loss_nll_class}
    \mathcal{L}_{\mathrm{NLL}}^{\mathrm{class}}
    =
    \mathbb{E}_{(x,u)}
    \left[
        -\log
        \left(
            \pi_{g(c)}(x)\,
            \phi_{g(c)}(u \mid x)
        \right)
    \right].
\end{equation}
This formulation encourages alignment between mixture components and known physical regimes while \textbf{preserving shared parameterization and statistical coupling} across components through a \textbf{common network architecture}. Joint training within a single model implicitly captures cross-component correlations induced by shared representations, which would generally be absent if each regime were modeled independently using separate networks.

Physics-based prior information is incorporated by augmenting the likelihood objective with additional constraint terms imposed on the component mean functions $\{\mu_m\}_{m=1}^M$. Since each mixture component represents a conditional mode with probability $\pi_m(x)$, physical consistency is enforced in expectation with respect to the latent component distribution. Let $\mathcal{R}_{\mathrm{phys}}(\mu_m;x)$ denote a physics-based residual or constraint functional acting on the $m$-th component mean. The physics-informed regularization term is defined as
\begin{equation}\label{eq:loss_phys}
    \mathcal{L}_{\mathrm{phys}}
    =
    \mathbb{E}_{(x)}
    \left[
        \sum_{m=1}^M
        \pi_m(x)\,
        \mathcal{R}_{\mathrm{phys}}\!\left(\mu_m; x\right)
    \right].
\end{equation}

The inclusion of the mixture weights $\pi_m(x)$ inside the expectation reflects the probabilistic semantics of mixture density models. In an MDN, each component corresponds to a latent conditional mode that is active with probability $\pi_m(x)$. As a result, physical consistency should not be enforced uniformly across all components, but rather in proportion to the likelihood that each component contributes to the conditional distribution at a given input.
Weighting the physics-based residual $\mathcal{R}_{\mathrm{phys}}(\mu_m; x)$ by $\pi_m(x)$ ensures that physical constraints are enforced in expectation with respect to the latent component distribution. Components that are unlikely under the data at a given input exert negligible influence on the regularization objective, whereas dominant components are constrained more strongly. This avoids imposing spurious physical penalties on inactive or weakly supported modes, which could otherwise bias the learned conditional distribution or distort multimodal structure.
From a probabilistic perspective, the regularization term in Equation~\ref{eq:loss_phys} corresponds to the expected physics residual under the joint distribution of inputs and latent mixture assignments. This formulation is consistent with the interpretation of MDNs as latent-variable models and enables physics-informed learning at the level of the full conditional distribution, rather than at the level of individual component realizations.

The total training objective is given by
\begin{equation}
    \mathcal{L}_{\mathrm{total}}
    =
    \mathcal{L}_{\mathrm{NLL}}
    +
    \lambda\,\mathcal{L}_{\mathrm{phys}},
\end{equation}
where $\lambda>0$ balances data fidelity and physical consistency.

\remark{
In discrete mixture models with fixed, sample-independent component
parameters, maximum likelihood estimation is commonly performed using the expectation--maximization (EM) algorithm, which alternates between estimating
latent component assignments and updating model parameters \cite{mclachlan1988mixture}.
In contrast, in the present setting the mixture coefficients and component parameters are modeled as deterministic, input-dependent functions parameterized by a single neural network. As a result, the likelihood can be optimized directly using gradient-based
methods without introducing an explicit EM procedure.
}

\subsection{Network Architecture}

The conditional mixture model is parameterized by a neural network that maps
the input variables $(z,x) \in \mathcal{Z} \times \mathcal{X}$ to the
parameters of the conditional mixture distribution.
Specifically, the network outputs the set
\[
\left\{ \pi_m(z,x), \mu_m(z,x), \sigma_m(z,x) \right\}_{m=1}^{M},
\]
corresponding to the mixture coefficients, component means, and component
standard deviations, respectively.

The mixture coefficients $\{\pi_m(z,x)\}_{m=1}^{M}$ are obtained by applying
a softmax activation to ensure non-negativity and normalization, such that
$\sum_{m=1}^{M} \pi_m(z,x) = 1$.
The component standard deviations
$\{\sigma_m(z,x)\}_{m=1}^{M}$ are enforced to be strictly positive through an
exponential mapping.
The component means $\{\mu_m(z,x)\}_{m=1}^{M}$ are produced without
additional constraints and serve as the primary carriers of physical structure
in the model.

\begin{figure}[htbp]
  \centering
  \includegraphics[width=0.8\textwidth]{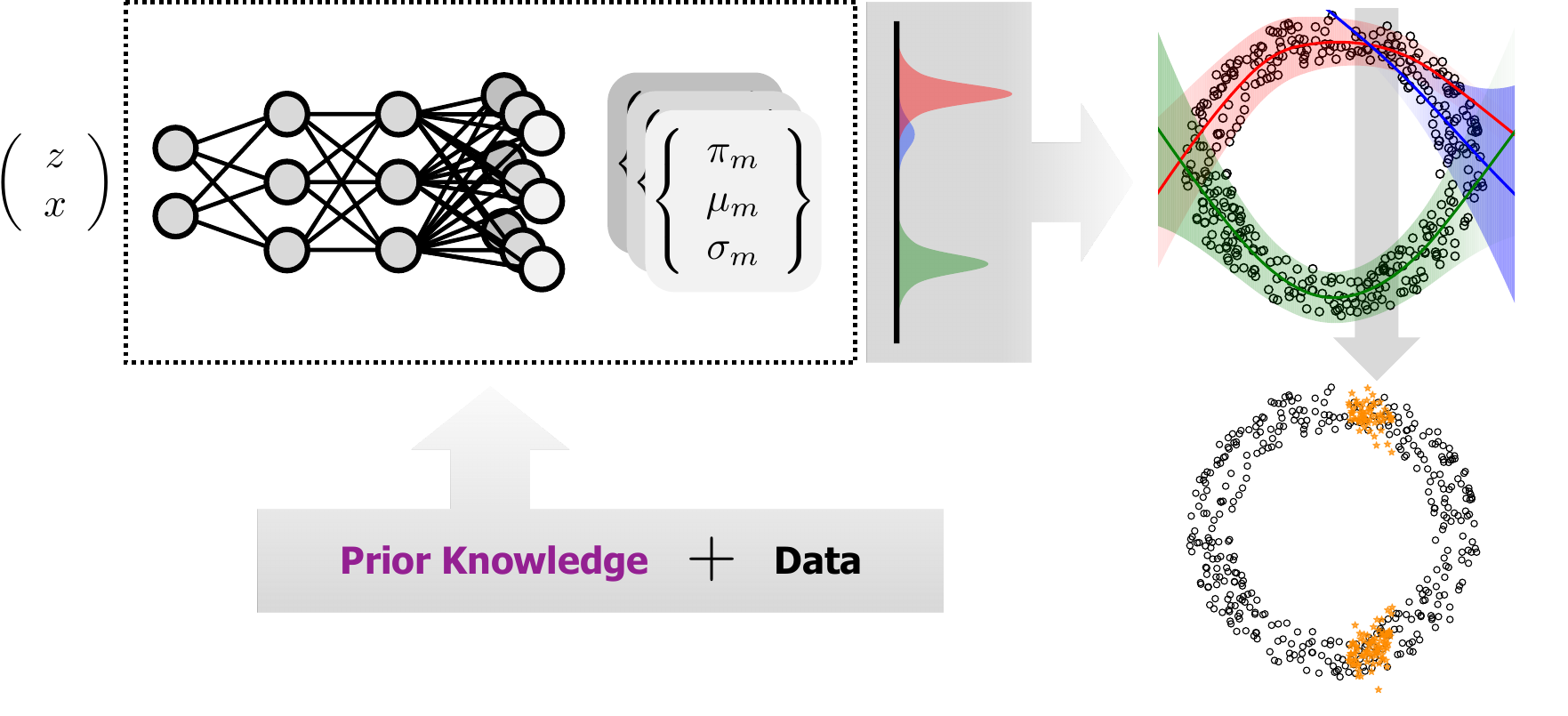}
  \caption{Illustrations of the MDN architecture and the proposed multimodal conditional modeling framework.}
  \label{fig:mdn_framework}
\end{figure}

Figure~\ref{fig:mdn_framework} illustrates the schematics of the MDN architecture and the proposed framework. 
A single network maps the input variables $(z,x)$ to the parameters of a conditional mixture distribution, including the mixture coefficients, means, and variances, thereby representing the full conditional density. The network architecture may consist of fully connected layers or other
appropriate architectures, depending on the structure of the input variables
$(z,x)$.
All mixture parameters are learned jointly via gradient-based optimizations.

\section{Numerical Examples}\label{sec:examples}

In this section, we present four numerical examples to illustrate the proposed framework.
We first consider a dynamical system exhibiting bifurcation behavior to compare the performance of MDN and CFM under comparable model capacity and training effort.
We then study a stochastic differential equation to evaluate the ability of MDNs to learn conditional equilibrium distributions and to assess their extrapolative behavior.
Next, we consider shock Hugoniot data aggregated from the literature and demonstrate how monotonicity constraints and class-conditional structure can be incorporated as prior knowledge within a single MDN to improve prediction accuracy and uncertainty estimates. Finally, we examine the Chafee--Infante equation to illustrate the effect of physics-informed regularization based on steady-state PDE residuals. Implementation details and hyperparameter settings are provided in the Appendix~\ref{appendix:hyperparams}.

Throughout all examples, automatic differentiation in \texttt{PyTorch} \cite{paszke2019pytorch} is used to evaluate the gradients required by physics-based loss terms.
While the number of mixture components must be specified \textit{a priori}, Appendix~\ref{appendix:circle} provides a supplementary study showing that the learned generative model exhibits limited sensitivity in both predictive performance and uncertainty estimates with respect to the number of mixture components.

\subsection{Bifurcation Diagram}\label{sec:sec:bifurcation}

For the first example, we consider a dynamical system exhibiting a cusp-type bifurcation,
\begin{equation}
    \dot{x} = -x^3 + \lambda x + \mu ,
    \label{eq:bifurcation_ode}
\end{equation}
where $x$ denotes the system state, $\lambda$ is a control parameter, and $\mu$ represents an imperfection parameter. This system has been used in the work of \cite{crabtree2025generative} as a benchmark for generative modeling of bifurcation surfaces and steady states in the joint $(x,\lambda,\mu)$ space. Here we treat this system as a conditional prediction problem under bounded imperfection $\mu$, and use it as a controlled test case to examine how mixture density models represent bifurcated, multimodal steady-state responses conditioned on the control parameter $\lambda$. We further compare the resulting mixture-based representation with a CFM generative model.

The system undergoes a symmetric pitchfork bifurcation for $\mu=0$, whereas nonzero $\mu$ breaks the symmetry and perturbs the bifurcation structure. We focus on the steady state response.

To construct the dataset, we sample $x \sim \mathcal{U}(-2,2)$ and $\lambda \sim \mathcal{U}(-2.5,2.5)$.
Only samples satisfying $|\mu| < 0.05$ are selected, corresponding to small but nonzero imperfections. This procedure yields data that are concentrated near the ideal bifurcation surface while incorporating small perturbations arising from imperfections.

\begin{figure}[htbp]
  \centering
  \begin{subfigure}{0.4\textwidth}
    \centering
    \includegraphics[width=\linewidth]{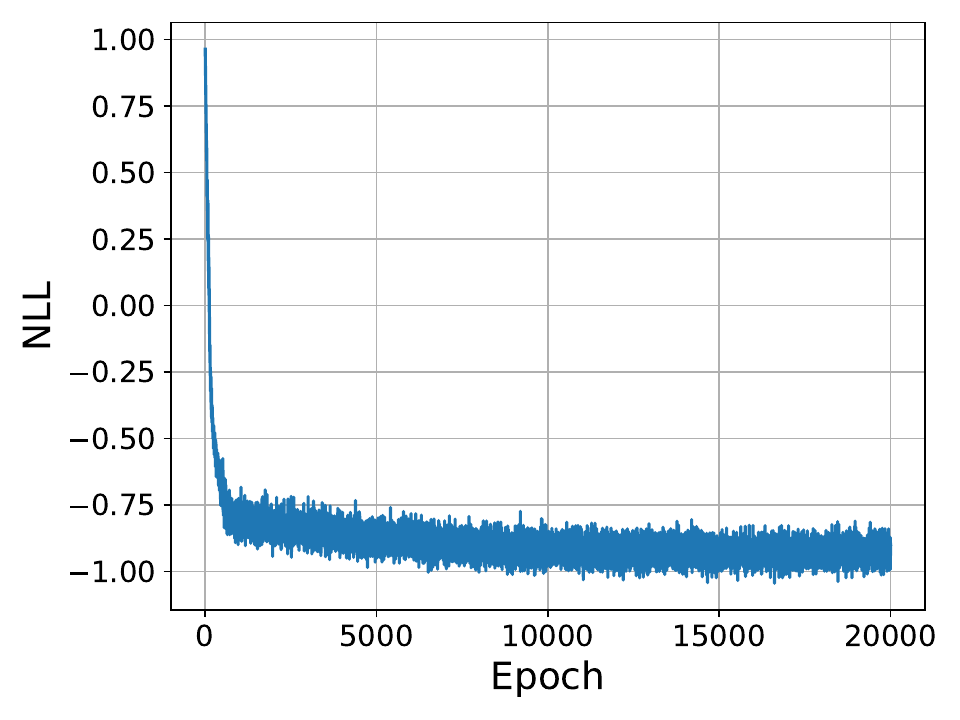}
    \caption{}
    \label{fig:bifurcation_main_a}
  \end{subfigure}
  \begin{subfigure}{0.4\textwidth}
    \centering
    \includegraphics[width=\linewidth]{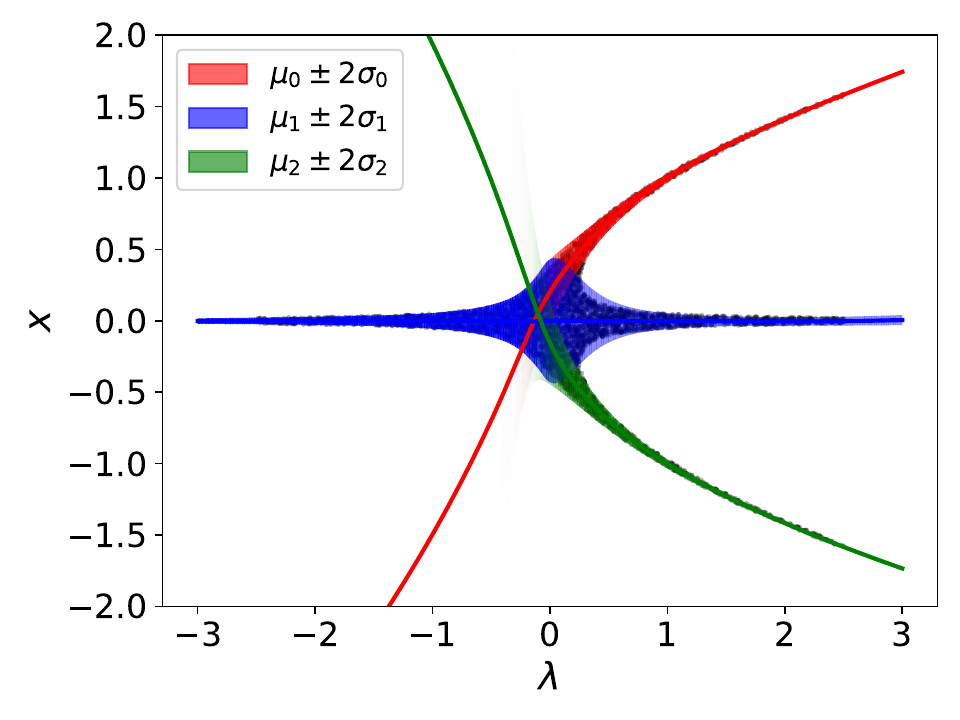}
    \caption{}
    \label{fig:bifurcation_main_b}
  \end{subfigure}
  \begin{subfigure}{0.4\textwidth}
    \centering
    \includegraphics[width=\linewidth]{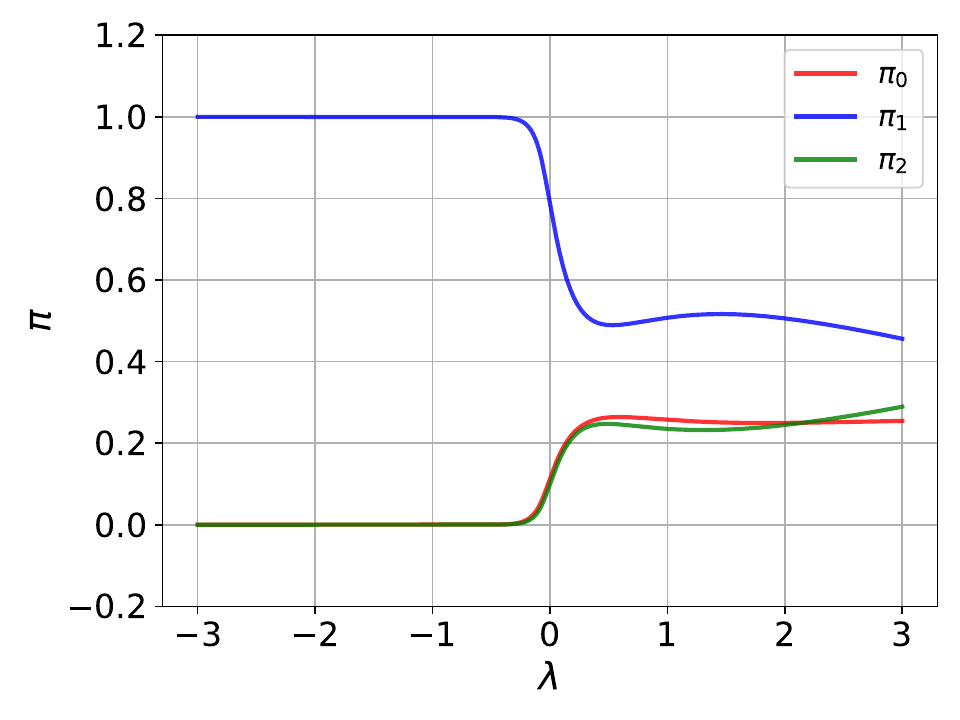}
    \caption{}
    \label{fig:bifurcation_main_c}
  \end{subfigure}
  \begin{subfigure}{0.4\textwidth}
    \centering
    \includegraphics[width=\linewidth]{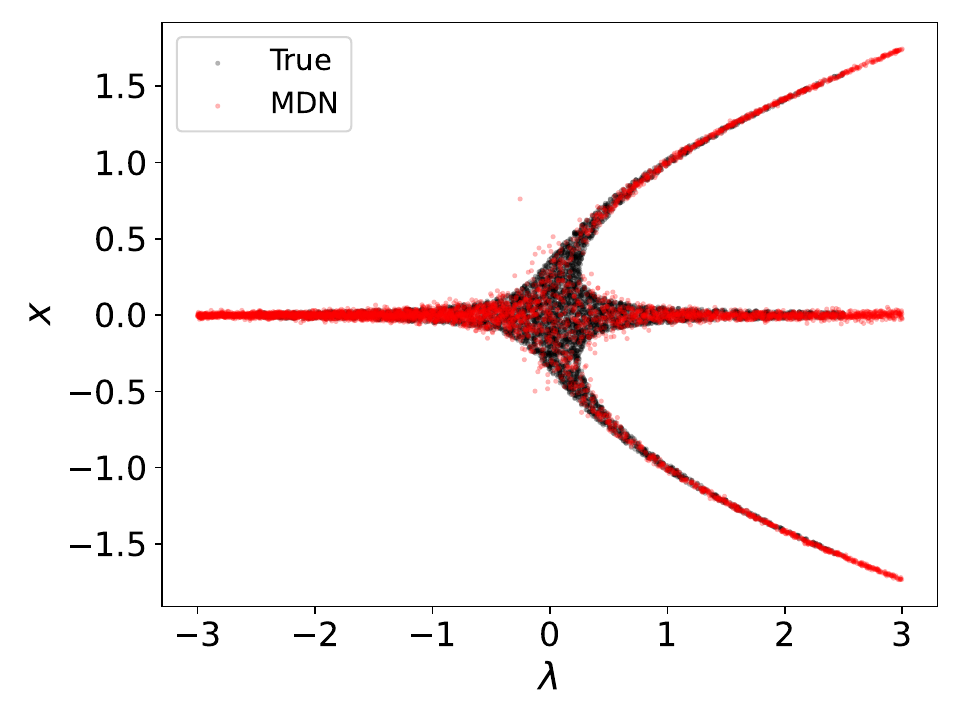}
    \caption{}
    \label{fig:bifurcation_main_d}
  \end{subfigure}
    \caption{
    MDN training for probabilistic modeling of bifurcated response branches.
    (\subref{fig:bifurcation_main_a}) Loss history over $20{,}000$ ADAM iterations.
    (\subref{fig:bifurcation_main_b}) Predicted component means $\mu_m$ and standard deviations $\sigma_m$ and  (\subref{fig:bifurcation_main_c}) predicted mixing coefficients $\pi_m$.
    (\subref{fig:bifurcation_main_d}) $5{,}000$ samples drawn from the trained MDN model.
    }
  \label{fig:bifurcation_main}
\end{figure}

For a fixed value of $\lambda$, multiple steady states may coexist due to the underlying bifurcation structure, while the bounded imperfection introduces variability around these admissible solutions.
As a result, as shown in Figure~\ref{fig:bifurcation_main}, the conditional distribution of $x$ given $\lambda$ is multimodal with nonzero spread.

The MDN training and the resulting probabilistic representation are illustrated in Figure~\ref{fig:bifurcation_main}.
 Figure~\ref{fig:bifurcation_main_a} shows the evolution of the negative log likelihood  objective function during optimization. 
      Figures~\ref{fig:bifurcation_main_b} and~\ref{fig:bifurcation_main_c} report the learned component-wise predicted mean fixed-point locations $\mu_m$, associated standard deviations $\sigma_m$, and mixing coefficients $\pi_m$ as functions of the conditioned parameter $\lambda$.
In regions $\lambda<0$ where a single steady state exists, the contribution of non-relevant components diminishes, whereas multiple components contribute in the bifurcated regime $\lambda>0$, reflecting the emergence of multimodal behavior.
Figure~\ref{fig:bifurcation_main_d} shows samples drawn from the trained MDN, illustrating that the model captures the bifurcated response structure together with the spread induced by bounded imperfection.

\begin{figure}[htbp]
  \centering
  \begin{subfigure}{0.4\textwidth}
    \centering
    \includegraphics[width=\linewidth]{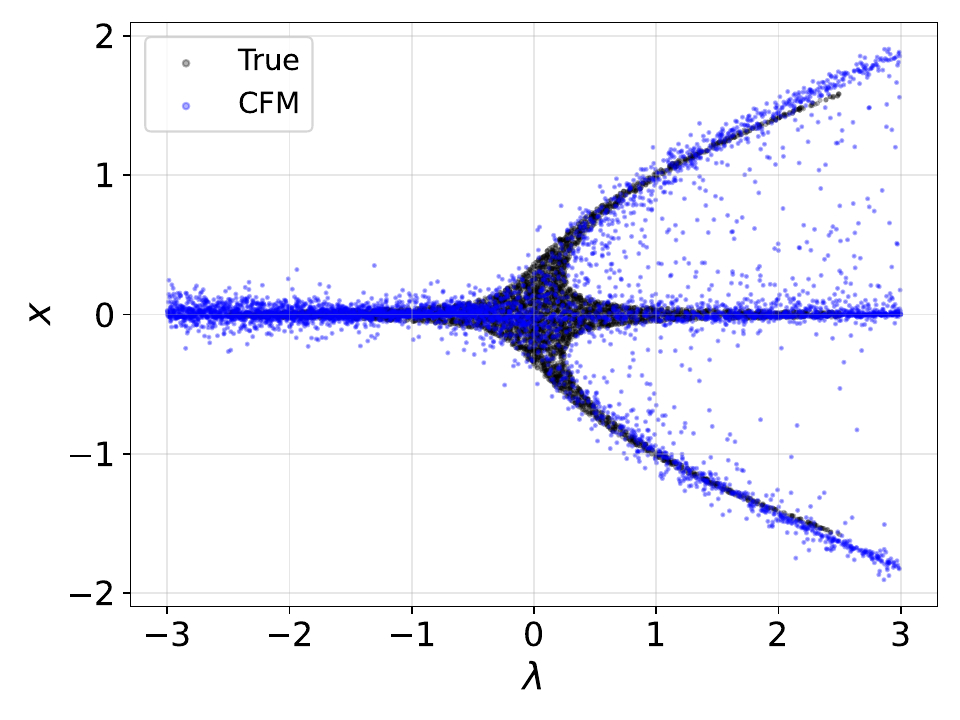}
    \caption{}
    \label{fig:bifurcation_sampling_a}
  \end{subfigure}
  \begin{subfigure}{0.4\textwidth}
    \centering
    \includegraphics[width=\linewidth]{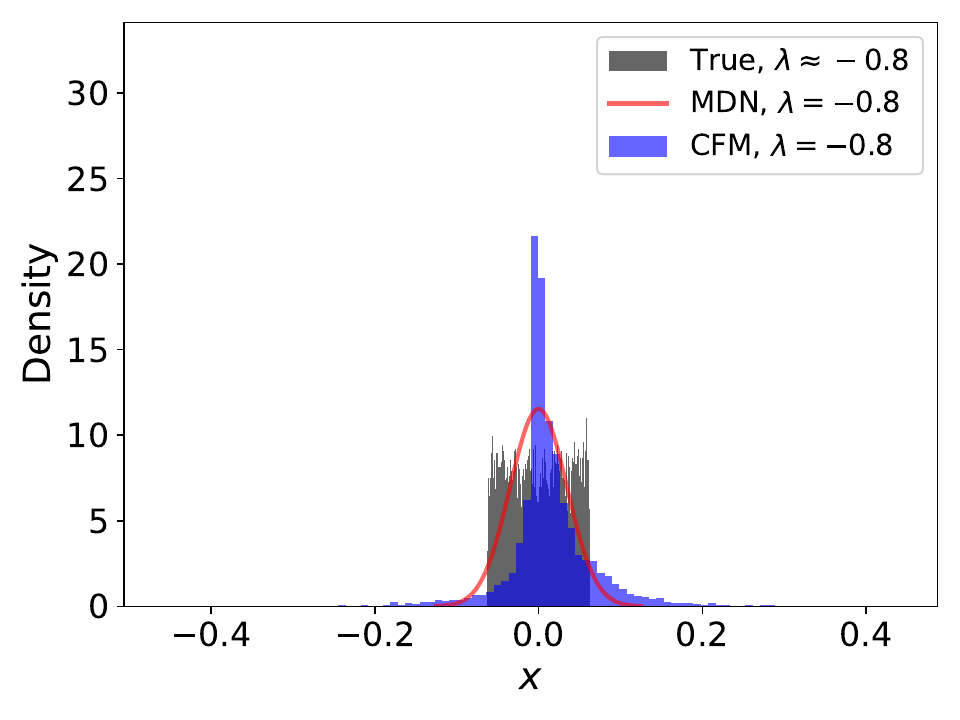}
    \caption{}
    \label{fig:bifurcation_sampling_b}
  \end{subfigure}
  \centering
  \begin{subfigure}{0.4\textwidth}
    \centering
    \includegraphics[width=\linewidth]{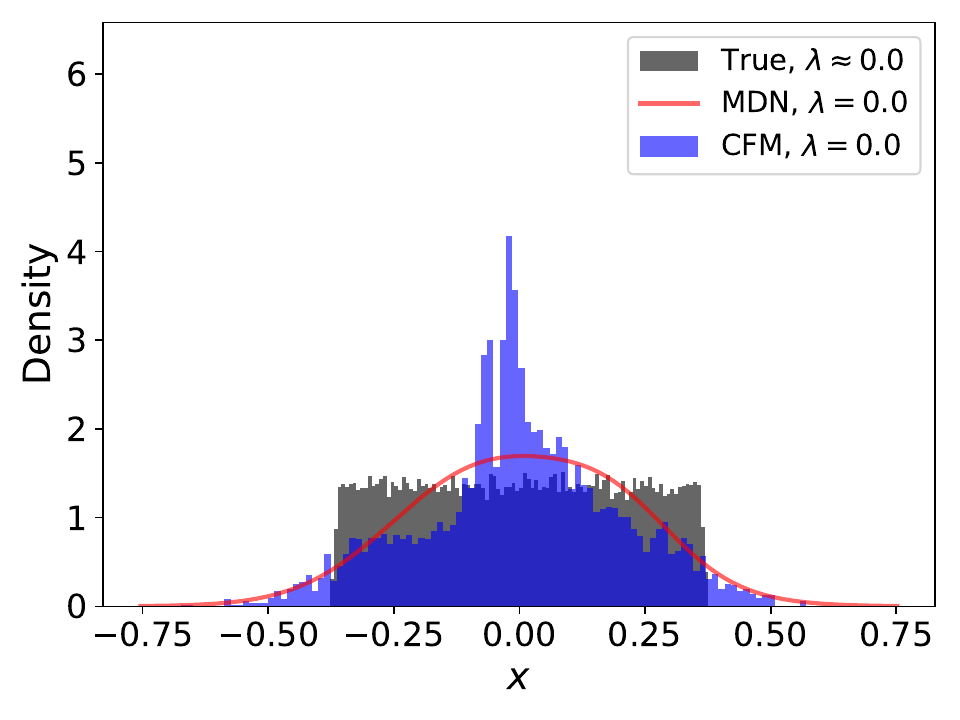}
    \caption{}
    \label{fig:bifurcation_sampling_c}
  \end{subfigure}
  \begin{subfigure}{0.4\textwidth}
    \centering
    \includegraphics[width=\linewidth]{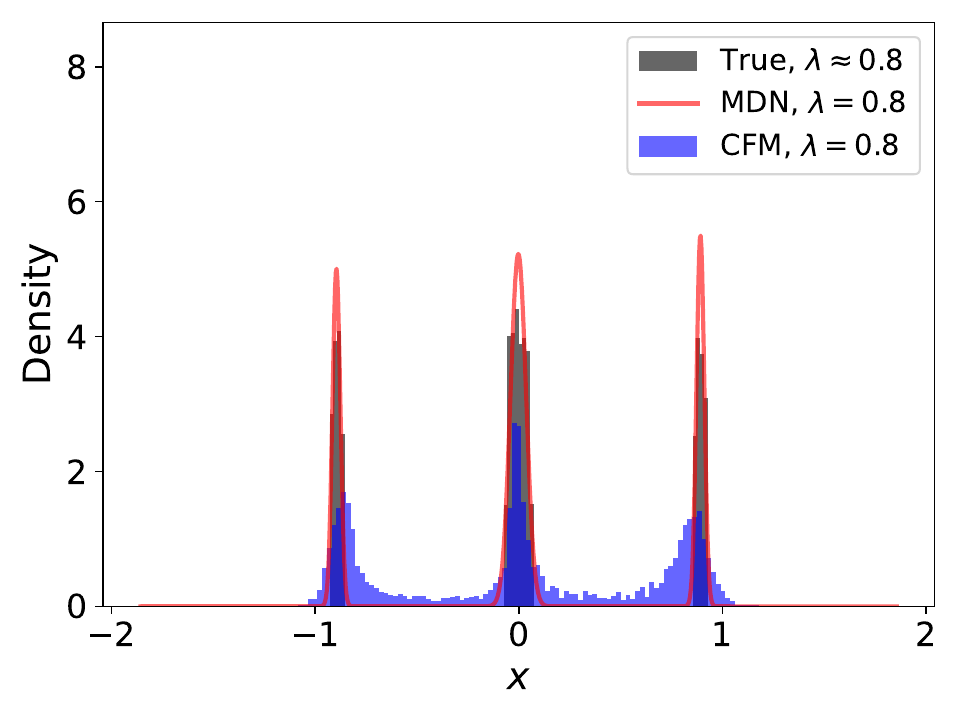}
    \caption{}
    \label{fig:bifurcation_sampling_d}
  \end{subfigure}
  \hfill
    \caption{
    Comparison between MDN and CFM both after $20{,}000$ ADAM iterations. The number of model parameters of MDN is $457$, where CFM consists of a comparable size of $521$ parameters.
    (\subref{fig:bifurcation_sampling_a}) Sample distribution of CFM model output conditioned on $5{,}000$ equispaced $\lambda$ values between $-3$ and $3$.
    Distribution of true distribution, MDN, and $5{,}000$ CFM samples, conditioned on
    (\subref{fig:bifurcation_sampling_b}) $\lambda=-0.8$,
    (\subref{fig:bifurcation_sampling_c}) $\lambda=0$, and
    (\subref{fig:bifurcation_sampling_d}) $\lambda=0.8$.
    }
  \label{fig:bifurcation_sampling}
\end{figure}

Figure~\ref{fig:bifurcation_sampling} compares the MDN and CFM models under comparable model capacity (similar number of trainable parameters).
Figure~\ref{fig:bifurcation_sampling_a} shows samples generated by the CFM conditioned on a dense set of control parameter values $\lambda \in [-3,3]$, illustrating that the model captures the overall bifurcation structure but exhibits noticeable dispersion across branches.
Figures~\ref{fig:bifurcation_sampling_b}–\subref{fig:bifurcation_sampling_d} compare the empirical conditional distributions of the true data, MDN, and CFM at representative conditioning values around $\lambda_0=-0.8$, $0$, and $0.8$.

For around $\lambda_0=-0.8$, where the response is unimodal, both models produce distributions concentrated near the single steady state.
At around $\lambda_0=0$, corresponding to the onset of bifurcation, the MDN captures the broadened conditional distribution, while the CFM exhibits a tendency to concentrate mass near intermediate values.
In the fully bifurcated regime $\lambda_0=0.8$, the MDN resolves three distinct response modes, whereas the CFM produces samples with noticeable probability mass in intermediate regions between modes.

These results illustrate that MDN can offer a competitive conditional generative modeling framework relative to state-of-the-art approaches such as CFM, while benefiting from substantially simpler training and sampling procedures. We emphasize that there may exist other settings in which CFM performs better than observed here; however, in this study, both methods were tuned with comparable effort and without extensive hyperparameter optimization.

\subsection{Multiscale Stochastic Differential Equations}\label{sec:sec:spde}

For the second example, we consider a two-dimensional multiscale stochastic differential equation,
\begin{align}\label{eq:spde}
    \mathrm{d}u_1 &= a_1\,\mathrm{d}t + a_2\,\mathrm{d}B_1,\\
    \mathrm{d}u_2 &= -\bigl(-1 + 0.2\,u_1 + 4\,u_2(-1 + u_2^2)\bigr)\,\mathrm{d}t
         + a_3\,\mathrm{d}B_2,
         \label{eq:spde_u2}
\end{align}
which has been studied in~\cite{crabtree2024micro} as a prototypical slow--fast stochastic system with parameter-dependent bistable behavior. In this system, the slow variable $u_1$ modulates the effective potential of the fast variable $u_2$, leading to a conditional equilibrium distribution of $u_2$ that is generally multimodal for fixed values of $u_1$.

Systems of this type provide a minimal yet representative setting for studying
conditional equilibrium measures in multiscale dynamics.
In many applications—such as molecular dynamics and materials modeling—one is
often interested in sampling fine-scale configurations consistent with a given
value of a coarse or collective variable.
In such settings, generative conditional models can play a complementary role
to physics-based simulations by providing informed initializations or proposal
distributions that accelerate sampling of the desired conditional measure.

In this example, we evaluate the performance of mixture density models in predicting the conditional distribution of $u_2$ given $u_1$. In addition, we examine the extrapolative behavior of the trained MDN beyond the range of conditioning values observed during training. This problem admits an analytic expression for the stationary conditional density, which is derived in Appendix~\ref{appendix:stationary_solution}.

\begin{figure}[htbp]
  \centering
  \begin{subfigure}{0.45\textwidth}
    \centering
    \includegraphics[width=\linewidth]{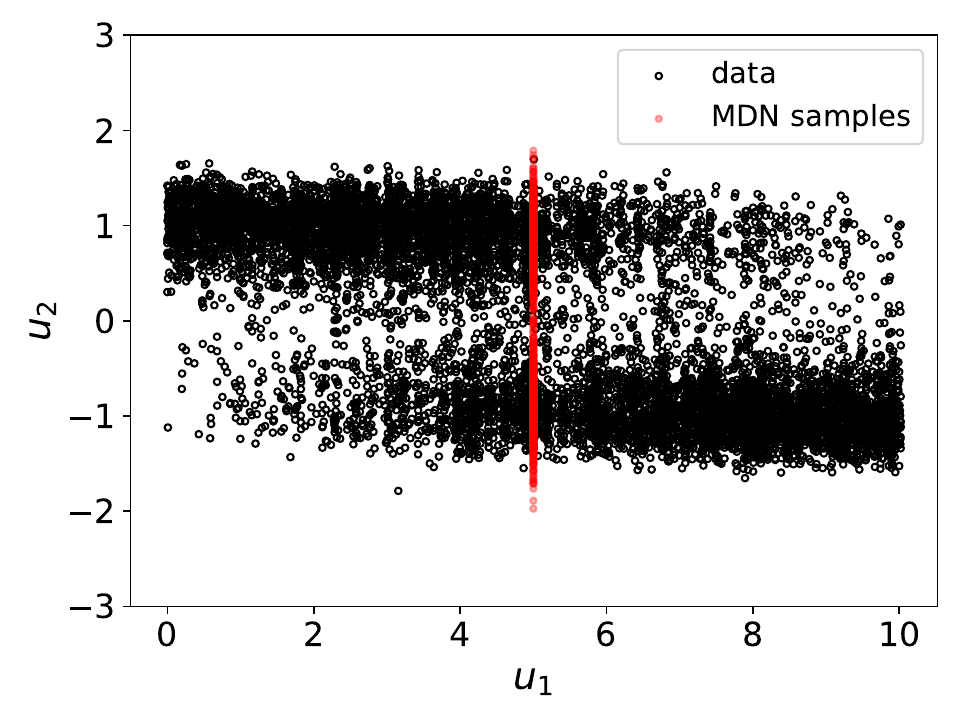}
    \caption{}
    \label{fig:spde_sampling_a}
  \end{subfigure}
  \begin{subfigure}{0.45\textwidth}
    \centering
    \includegraphics[width=\linewidth]{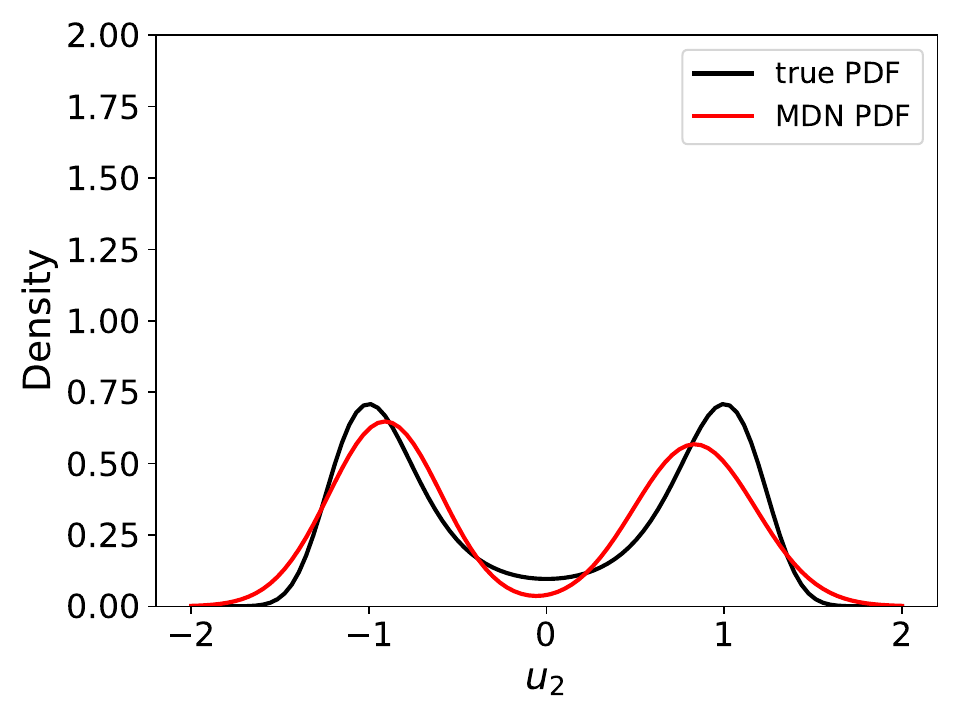}
    \caption{}
    \label{fig:spde_sampling_b}
  \end{subfigure}
  \centering
  \begin{subfigure}{0.45\textwidth}
    \centering
    \includegraphics[width=\linewidth]{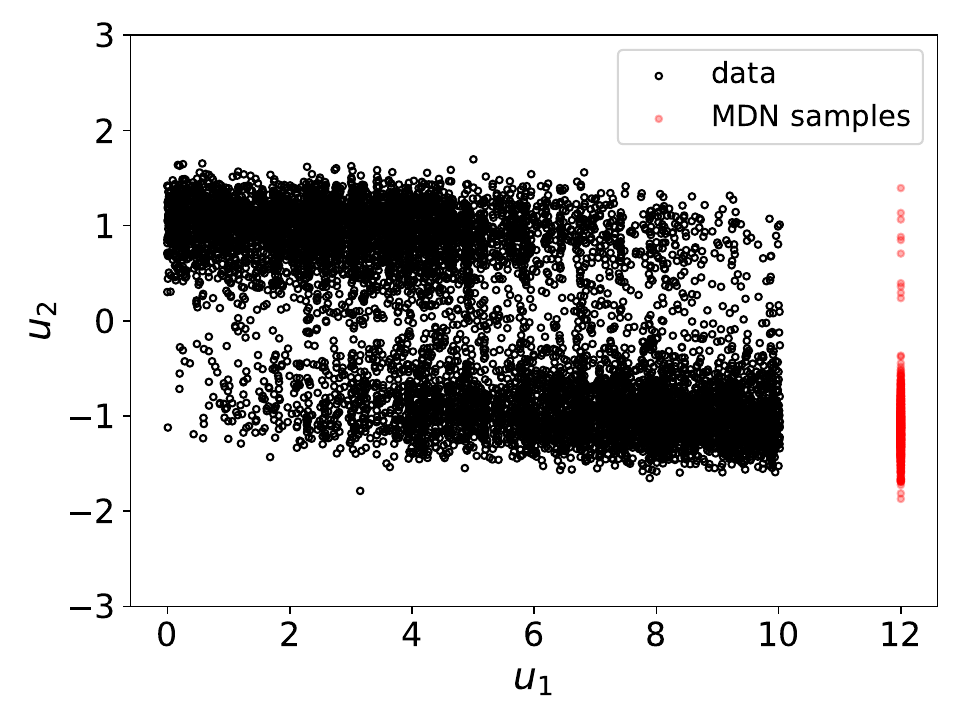}
    \caption{}
    \label{fig:spde_sampling_c}
  \end{subfigure}
  \begin{subfigure}{0.45\textwidth}
    \centering
    \includegraphics[width=\linewidth]{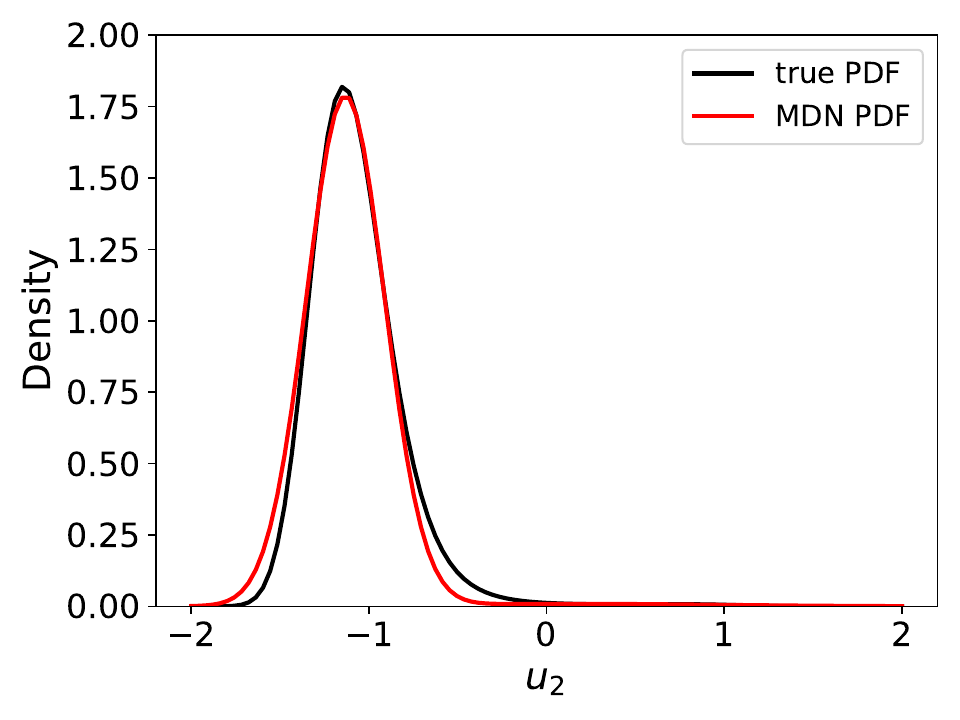}
    \caption{}
    \label{fig:spde_sampling_d}
  \end{subfigure}
  \hfill
    \caption{
    Results of training MDNs with two mixture components on SPDE data. SPDE data are randomly subsampled into size $10{,}000$ out of $10{,}000{,}000$ data points for visibility.
    (\subref{fig:spde_sampling_a}) $1{,}000$ samples from MDN model conditioned on $u_1=5$, and 
    (\subref{fig:spde_sampling_b}) corresponding PDF of MDN and true distribution.
    (\subref{fig:spde_sampling_c}) $1{,}000$ samples from MDN model conditioned on $u_1=12$, and 
    (\subref{fig:spde_sampling_d}) corresponding PDF of MDN and true distribution.
    }
  \label{fig:spde_sampling}
\end{figure}

Figure~\ref{fig:spde_sampling} illustrates the performance of the MDN on the multiscale SDE example. 
Figures~\ref{fig:spde_sampling_a} and~\ref{fig:spde_sampling_b} show MDN outputs queried at $u_1=5$. Figure~\ref{fig:spde_sampling_a} displays samples generated by the MDN conditioned on this value, while Figure~\ref{fig:spde_sampling_b} compares the analytic stationary conditional density with the density predicted by the MDN, showing that the bimodal structure of the conditional distribution is captured.
Figures~\ref{fig:spde_sampling_c} and~\ref{fig:spde_sampling_d} report the same quantities for $u_1=12$, which lies outside the training range. Despite this extrapolative setting, the MDN continues to produce samples consistent with the data and recovers the overall structure of the true conditional density. These results indicate that the MDN accurately captures the multimodal conditional distribution of the fast variable and remains effective when queried beyond the range of conditioning values seen during training.

\subsection{Shockwave Physics}\label{sec:sec:shockwave}

\begin{figure}[htbp]
  \centering
  \begin{subfigure}{0.6\textwidth}
    \centering
    \includegraphics[width=\linewidth]{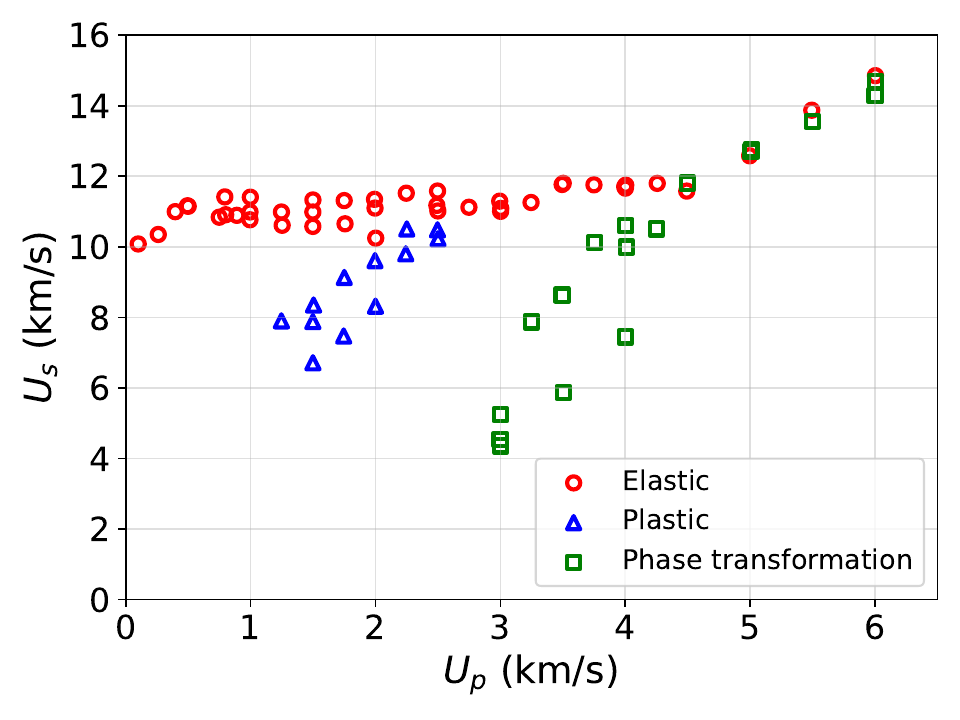}
  \end{subfigure}
  \hfill
    \caption{
    Shock Hugoniots of single-crystal 3C--SiC data along the $[001]$ orientation, digitized and reproduced from \citep{li2017shock,branicio2018plane,pasparakis2026physics}.
    }
  \label{fig:shockwave_data}
\end{figure}

In third example, we consider shock Hugoniots of single-crystal 3C--SiC along the $[001]$ orientation, expressed in terms of the shock velocity--particle velocity ($U_s$--$U_p$) relation.
The aggregated dataset spans multiple deformation regimes, including elastic response, plastic flow, and shock-induced phase transformation, reflecting the evolution of wave structure under increasing shock strength.
The $U_s$–$U_p$ data are compiled from molecular dynamics simulations and experimental comparisons reported in~\citep{pasparakis2026physics,li2017shock,branicio2018plane}, as shown in Figure~\ref{fig:shockwave_data}. We reproduced data from the published figures using plot digitization. Obtaining each of these data points requires costly computational simulations (e.g., large-scale molecular dynamics simulations) or experiments, which motivates the use of machine learning models as surrogates. More importantly, probabilistic surrogate models can guide subsequent experiments or simulations by identifying informative regimes for further exploration.

The elastic, plastic, and phase-transformation regimes occupy distinct branches in the $U_s$–$U_p$ plane, reflecting different response mechanisms and multimodal structure. Consistent with both the data and thermodynamic considerations, the shock velocity $U_s$ is a non-decreasing function of the particle velocity $U_p$. We incorporate this property as a physics-informed prior by enforcing a monotonicity constraint through a physics-based regularization term:
\begin{equation}\label{eq:monoton_loss}
    \mathcal{R}_{\text{phys}}(\mu_m;x)=
        \max\left(0,-\partial \mu_m\right).
\end{equation}

\begin{figure}[htbp]
  \centering
  \begin{subfigure}{0.4\textwidth}
    \centering
    \includegraphics[width=\linewidth]{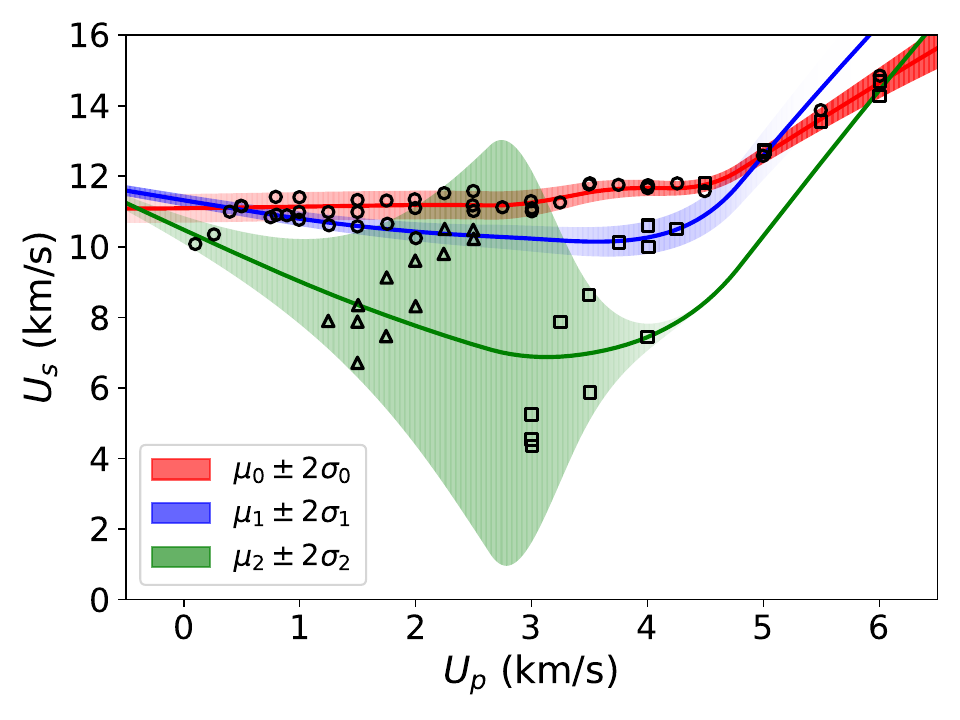}
  \end{subfigure}
  \begin{subfigure}{0.4\textwidth}
    \centering
    \includegraphics[width=\linewidth]{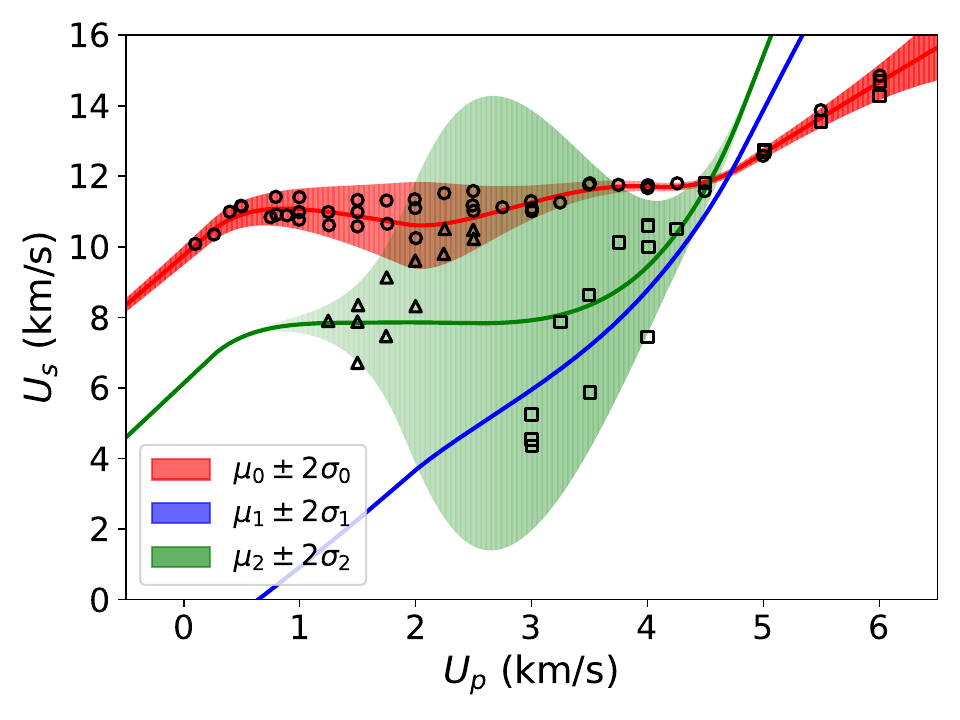}
  \end{subfigure}
  \begin{subfigure}{0.4\textwidth}
    \centering
    \includegraphics[width=\linewidth]{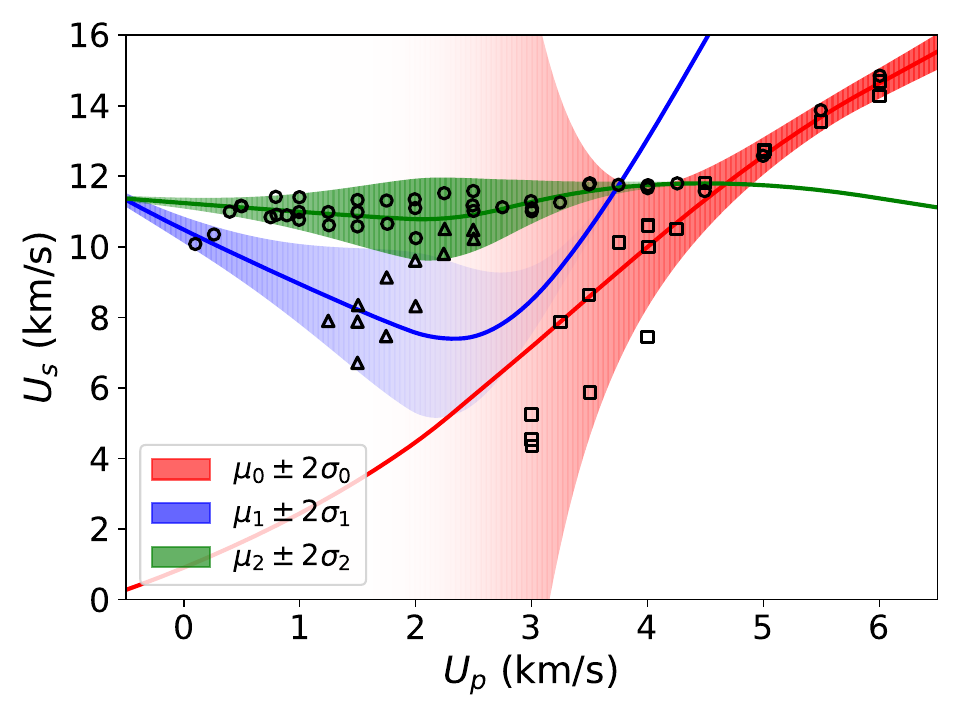}
  \end{subfigure}
  \begin{subfigure}{0.4\textwidth}
    \centering
    \includegraphics[width=\linewidth]{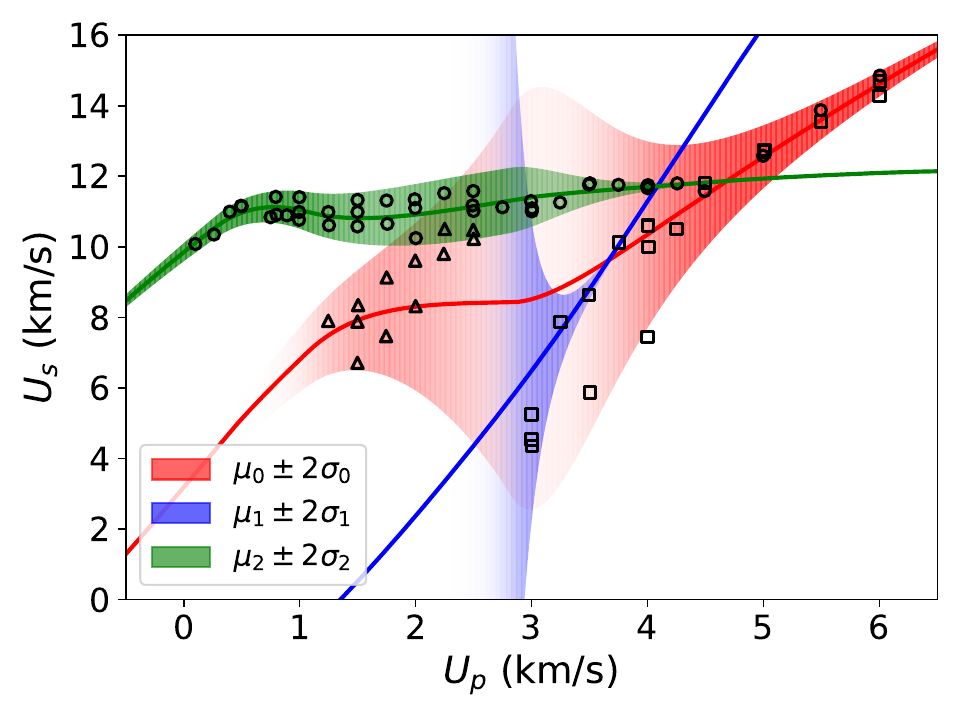}
  \end{subfigure}
  \begin{subfigure}{0.4\textwidth}
    \centering
    \includegraphics[width=\linewidth]{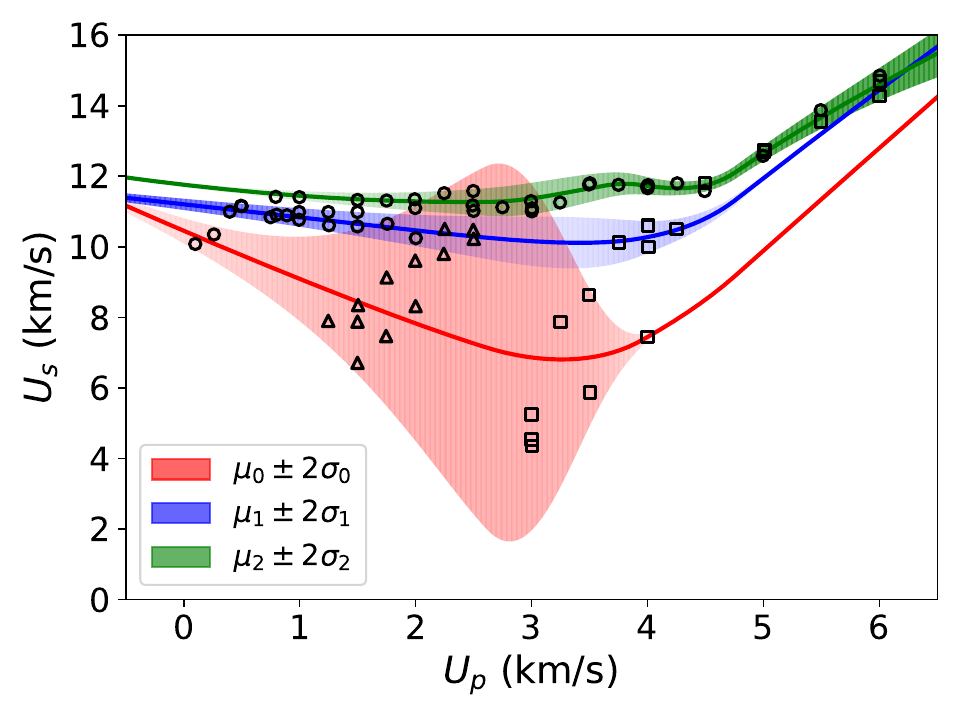}
  \end{subfigure}
  \begin{subfigure}{0.4\textwidth}
    \centering
    \includegraphics[width=\linewidth]{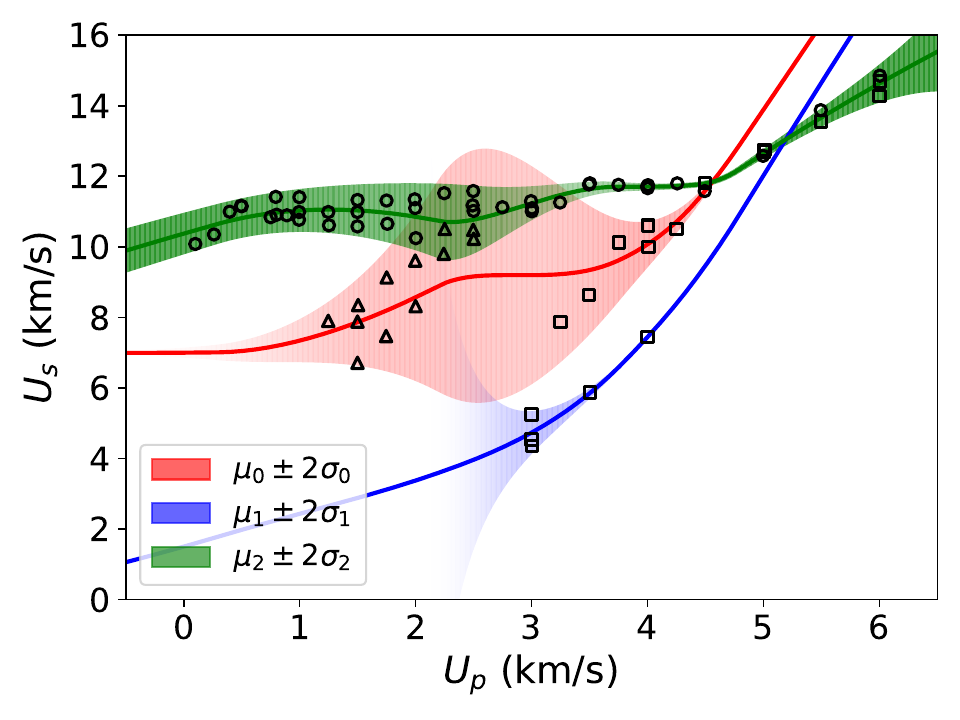}
  \end{subfigure}
  \hfill
    \caption{
    Effect of prior knowledge on training MDN model.
    Results of three different trials after $10{,}000$ ADAM iterations, (left) only with NLL loss and (right) combined with prior knowledge of monotone increasing.
    Each marker refers to different regimes: elastic as circles, plastic as triangles, and phase transformation as squares.
    }
  \label{fig:shockwave_monoton}
\end{figure}

Figure~\ref{fig:shockwave_monoton} illustrates the effect of incorporating monotonicity as prior physical knowledge in MDN training. The left column shows results from three independent trials trained using only the NLL loss Equation~\ref{eq:loss_nll}, while the right column shows the corresponding trials trained with the additional monotonicity regularization term Equation~\ref{eq:monoton_loss}.
When trained with the NLL loss alone, the learned component means exhibit noticeable variability across trials and do not consistently respect the expected monotone increasing relationship between $U_s$ and $U_p$. In particular, the mean predictions tend to prioritize fitting the overall data distribution, leading to non-monotone behavior in intermediate regions. By contrast, when the monotonicity prior is included, the learned component means more consistently follow a physical trend with respect to $U_p$ across trials. While this weak physical constraint does not fully separate the elastic, plastic, and phase-transformation regimes, it mitigates nonphysical behavior observed in the unconstrained setting and yields predictions that better reflect the underlying shock physics.

\begin{figure}[htbp]
  \centering
  \begin{subfigure}{0.32\textwidth}\centering
    \includegraphics[width=\linewidth]{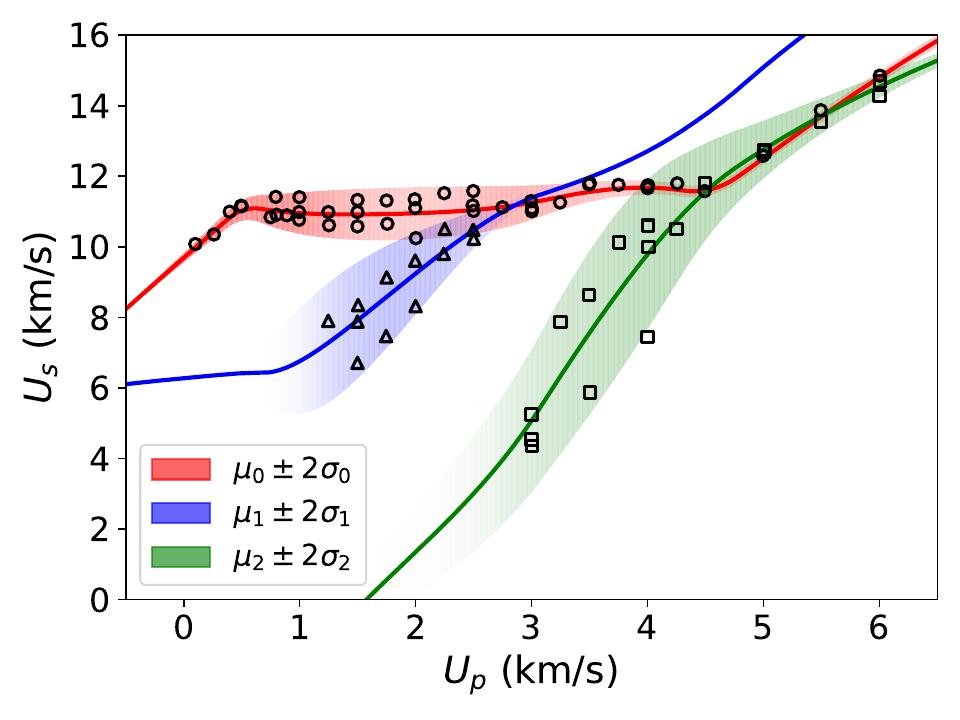}
  \end{subfigure}
  \begin{subfigure}{0.32\textwidth}\centering
    \includegraphics[width=\linewidth]{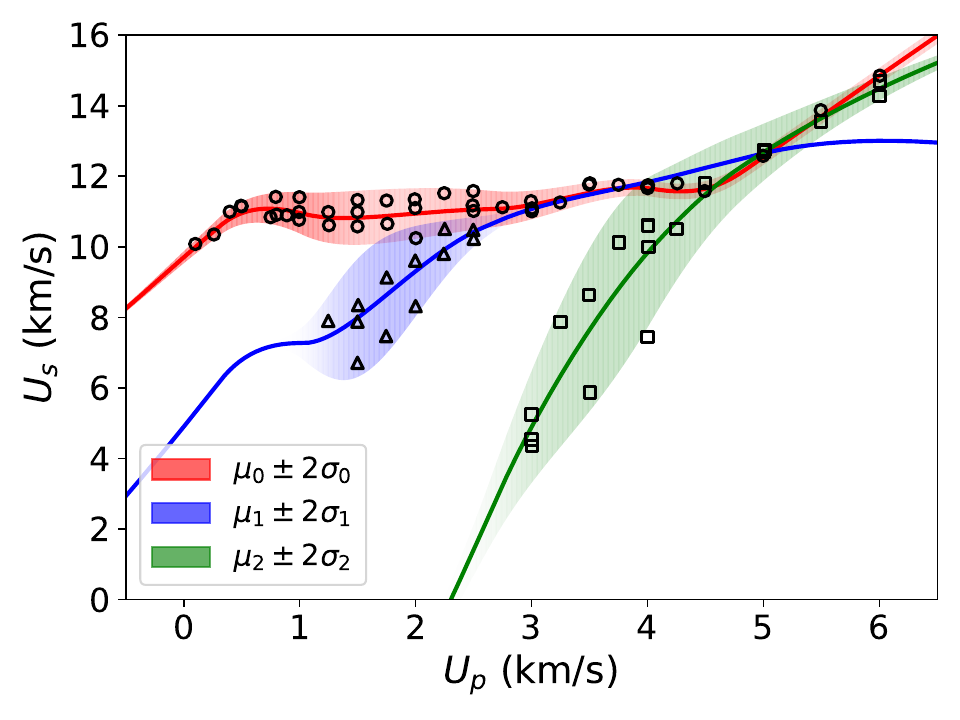}
  \end{subfigure}
  \begin{subfigure}{0.32\textwidth}\centering
    \includegraphics[width=\linewidth]{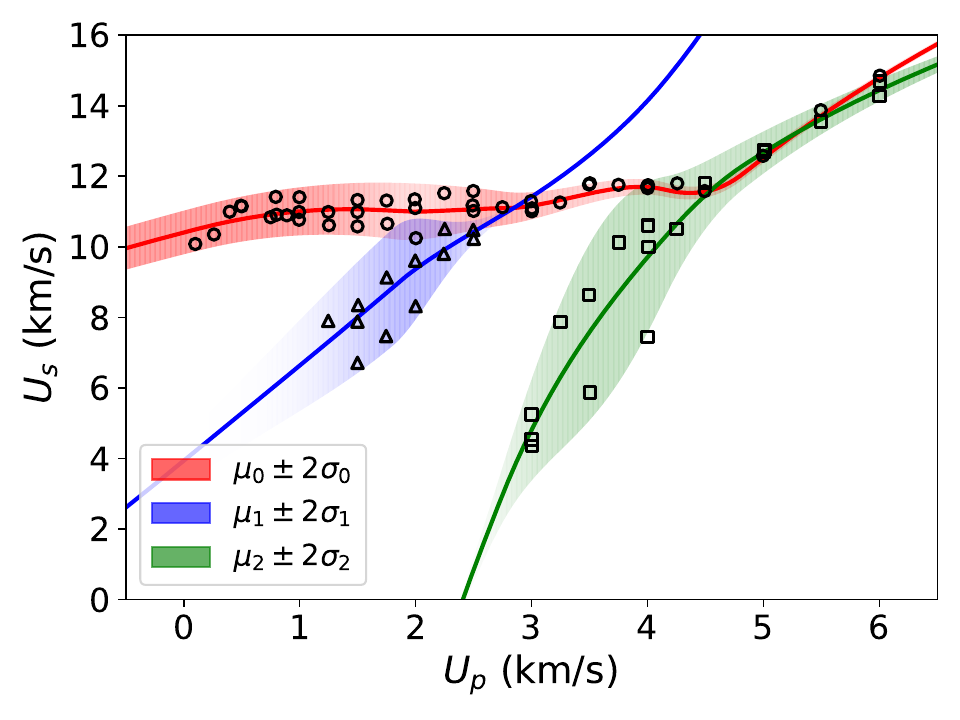}
  \end{subfigure}
  \hfill
    \caption{
    Three different trials of training MDN models with the prior knowledge of class conditional by one-hot-encoding and monotone increasing, over $10{,}000$ ADAM iterations.
    }
  \label{fig:shockwave_phys}
\end{figure}

In this problem, class labels are known \emph{a priori} and can be leveraged as additional supervision to improve modeling performance.  
In this setting, each data point is labeled according to its regime, and mixture components are explicitly assigned to these regimes.
Figure~\ref{fig:shockwave_phys} shows the effect of incorporating both class-conditional structure Equation~\ref{eq:loss_nll_class} and monotonicity Equation~\ref{eq:monoton_loss} as prior knowledge in MDN training.
Compared with the non-class-informed results in Figure~\ref{fig:shockwave_monoton}, the class-informed model yields a clearer separation of response branches, with component means that are more consistently aligned with their corresponding regimes.
These results highlight the utility of MDNs as flexible surrogates for multimodal physical systems, capable of integrating data-driven learning with interpretable, physics-aligned structure.

The variability across different model initializations persists even with the incorporation of physics-based knowledge. This behavior is primarily due to the data-limited nature of the problem, with most variability occurring in regions lacking data support. Moreover, the present formulation does not explicitly model epistemic uncertainty.

\subsection{Differential Equation with Random Field Initial Condition}\label{sec:sec:chafee}

For the last example, we consider a Chafee--Infante reaction--diffusion equation
\begin{align}
    &u_t = u - u^3 + \nu \Delta u, \quad x \in [0,\pi], \; t \in [0,\infty), \\
    &u(0,t) = u(\pi,t) = 0.
    \label{eq:chafee_pde}
\end{align}
This problem has previously been studied in \cite{crabtree2025generative} as a benchmark for a conditional score-based generative model \cite{song2020score}. Here, we choose \(\nu = 0.16\), for which the long-time dynamics admit multiple stable solutions and are known to evolve on a two-dimensional inertial manifold \cite{gear2011slow}.
%
The simulations are initialized from random initial conditions of the form
\begin{equation}
u(x,0) = \sum_{n=1}^{3} a_n \sin(nx),
\end{equation}
where $a_n \sim \mathcal{N}(0,1)$.
100 solution profiles at the time $t=4.5$ (close to steady state) are collected for training.

The physics-informed loss in this problem is defined based on the steady-state PDE residual as follow,
\begin{equation}\label{eq:chafee_residual}
    \mathcal{R}_{\text{phys}}(\mu_m;x)=
    \left(
        \mu_m-\mu_m^3+\nu\Delta\mu_m
    \right)^2.
\end{equation}

\begin{figure}[htbp]
  \centering
  \begin{subfigure}{0.4\textwidth}
    \centering
    \includegraphics[width=\linewidth]{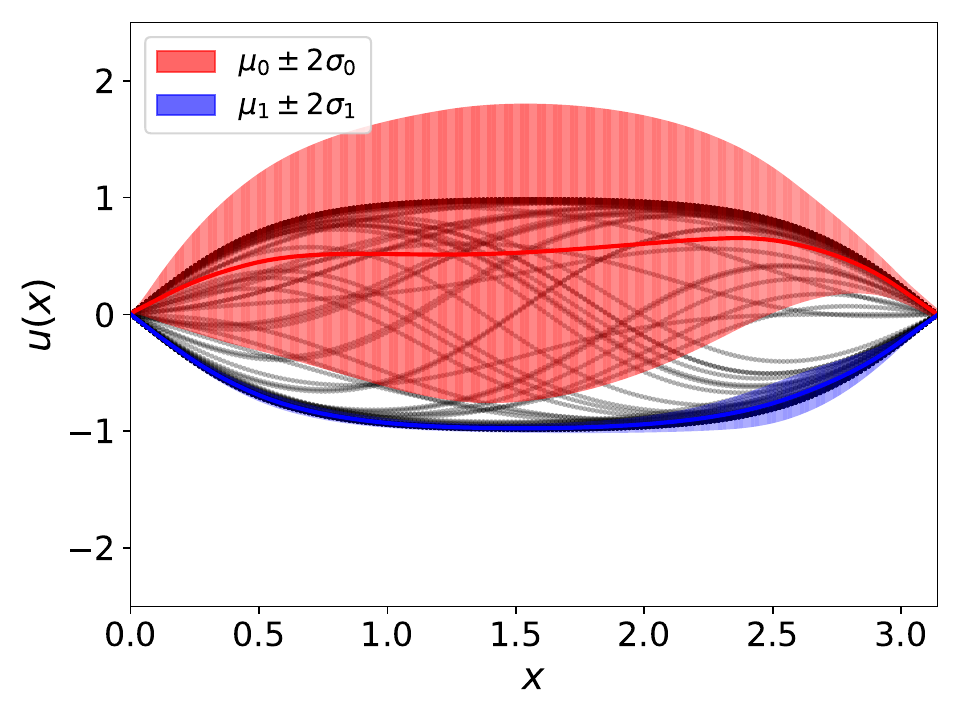}
    \caption{}
    \label{fig:chafee_loss_scheme_a}
  \end{subfigure}
  \begin{subfigure}{0.4\textwidth}
    \centering
    \includegraphics[width=\linewidth]{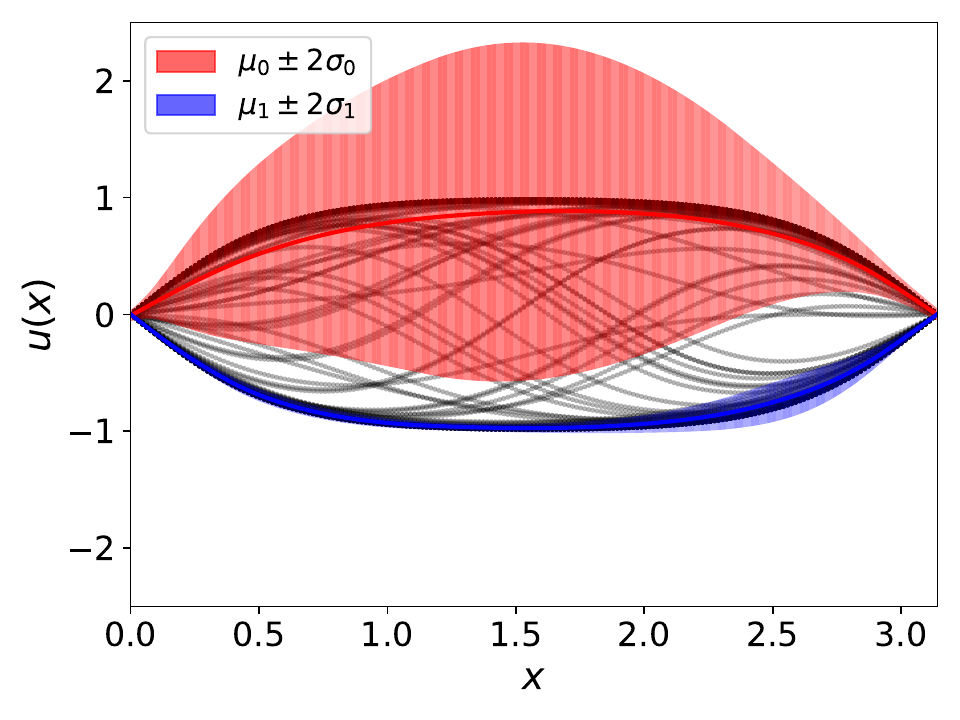}
    \caption{}
    \label{fig:chafee_loss_scheme_b}
  \end{subfigure}
  \hfill
    \caption{
    Comparison between the MDN models with 
    (\subref{fig:chafee_loss_scheme_a}) a vanilla NLL loss and 
    (\subref{fig:chafee_loss_scheme_b}) a combination of NLL and physics-based prior, after $50{,}000$ ADAM iterations.
    }
  \label{fig:chafee_loss_scheme}
\end{figure}

Figure~\ref{fig:chafee_loss_scheme} compares MDN predictions obtained under different training objectives. As shown, the data concentrate around two distinct solution profiles. Figure~\ref{fig:chafee_loss_scheme_a} shows the results obtained by training the MDN using only the NLL loss.
Statistically, the solution ensemble concentrates predominantly around the two stable modes (up and down), which the model is able to roughly distinguish. However, one of the predicted mixture means is noticeably displaced from the corresponding steady-state solution. This discrepancy is due to the presence of intermediate transient states in the training data that have not yet converged to equilibrium. We intentionally include such transient trajectories in the dataset to assess the model’s performance.

Figure~\ref{fig:chafee_loss_scheme_b} shows the result obtained by augmenting the NLL loss with the physics-informed loss. Once the physics-based prior is imposed, the model becomes biased toward mean solutions. As a result, the predicted mixture means are closer to the stable steady-state solutions compared to the previous case. At the same time, the predicted standard deviations increase. This behavior is expected, as the mixture means are constrained toward physically admissible steady states, the model compensates by increasing the variance to account for residual data variability arising from non-equilibrated states. Although these states carry less probability mass than the stable solutions, their contribution remains nonzero, and the increased variance enables the model to represent them in a statistically consistent manner.

The symmetric Gaussian standard deviations predicted in this example may appear unrealistic, since the data do not populate one side of the stable solutions. This is a consequence of the imposed conditional Gaussian assumption in the MDN. Extensions to more flexible conditional distributions could address this behavior, but are not considered here.

\section{Conclusion}\label{sec:conclusion}

In this work, we study physics-constrained multimodal conditional modeling using MDN as a flexible and interpretable probabilistic framework.
Across a range of representative examples—including bifurcation phenomena, multiscale stochastic dynamics, atomistic-scale shock physics, and reaction--diffusion equations—we illustrate the use of MDN for modeling multimodal conditional distributions arising from intrinsic physical mechanisms.

We further show that diverse forms of prior knowledge—including thermodynamic principles, PDE-based constraints, and class-conditional structure—can be seamlessly integrated into the MDN within the proposed framework. In particular, we introduce a class-conditional formulation that is enforced as a regularizer during training, guiding the learned mixture components toward physically and semantically meaningful partitions of the conditional distribution.
The incorporation of such prior knowledge enhances component-wise agreement with known physical behavior without requiring additional data or the introduction of separate class-specific models, thereby preserving shared structure and statistical coupling across mixture components within a single probabilistic model.

Moreover, we demonstrate that MDN can achieve performance comparable to, and in some settings exhibit more structured behavior than, CFM (a state-of-the-art model) under similar model capacity and training effort, while retaining the practical advantages of simpler training and sampling procedures enabled by the explicit mixture representation. 

Importantly, the explicit mixture structure yields enhanced interpretability: the conditional probability space is decomposed into a finite set of components corresponding to \textit{a priori} defined regions or regimes. This decomposition enables targeted diagnosis and post-processing of the generative mechanism by focusing on individual components with reduced model complexity. Furthermore, because the conditional distribution is available in closed form, uncertainty quantification becomes more transparent and analytically tractable—allowing direct computation of first and second conditional statistical moments without reliance on Monte Carlo sampling.

\textbf{Limitation.} While the conditional Gaussian assumption is appropriate for many problems, including several considered in this study, it may be unrealistic in others, such as the final example. Although the proposed framework accommodates more expressive conditional distributions, these extensions are not explored here. Furthermore, while the model captures predictive uncertainty, it does not explicitly account for model uncertainty.

\section*{Data Availability}
All data used in this work are either publicly available or can be reproduced from the information provided and are available from the corresponding author upon reasonable request.

\section*{Acknowledgment}
B.B. acknowledges fruitful discussions with Michael Shields and Yannis Kevrekidis during Spring 2025, particularly regarding the shock physics dataset and score-based generative modeling. B.B. also acknowledges support from the startup fund provided by Northwestern University.

\section*{Declaration on the Use of Generative AI}

The authors used ChatGPT, a large language model developed by OpenAI, to assist with English language editing and grammar refinement. The scientific content, technical interpretations, and conclusions are solely the responsibility of the authors.
\bibliographystyle{plainnat}
\bibliography{bibliography}

\appendix

\section{Network Architecture}\label{appendix:hyperparams}

Hyperparameter settings for each problem are summarized below, following the order of presentation in Section~\ref{sec:examples} and Appendix~\ref{appendix:circle}.
All models are trained using a learning rate of $10^{-3}$, two hidden layers, and ELU activation functions.
The number of hidden units is set to $16$ for the bifurcation example with MDN, $20$ for the bifurcation example with CFM, and $32$ for the multiscale SDE, shockwave, Chafee--Infante, and circle examples.

\section{Circle Data}\label{appendix:circle}

As a supplementary example, we consider a synthetic two-dimensional circular dataset with conditional multimodality. Points are generated by sampling angles uniformly in $[0,2\pi)$ and radii with density proportional to area within an annulus of inner radius $r_{\mathrm{in}}=0.35$ and outer radius $r_{\mathrm{out}}=0.5$, centered at $(0.5,0.5)$. A total of $400$ data points are used. Results obtained with different numbers of mixture components are compared under identical training settings.

\begin{figure}[htbp]
  \centering
  \begin{subfigure}{0.32\textwidth}
    \centering
    \includegraphics[width=\linewidth]{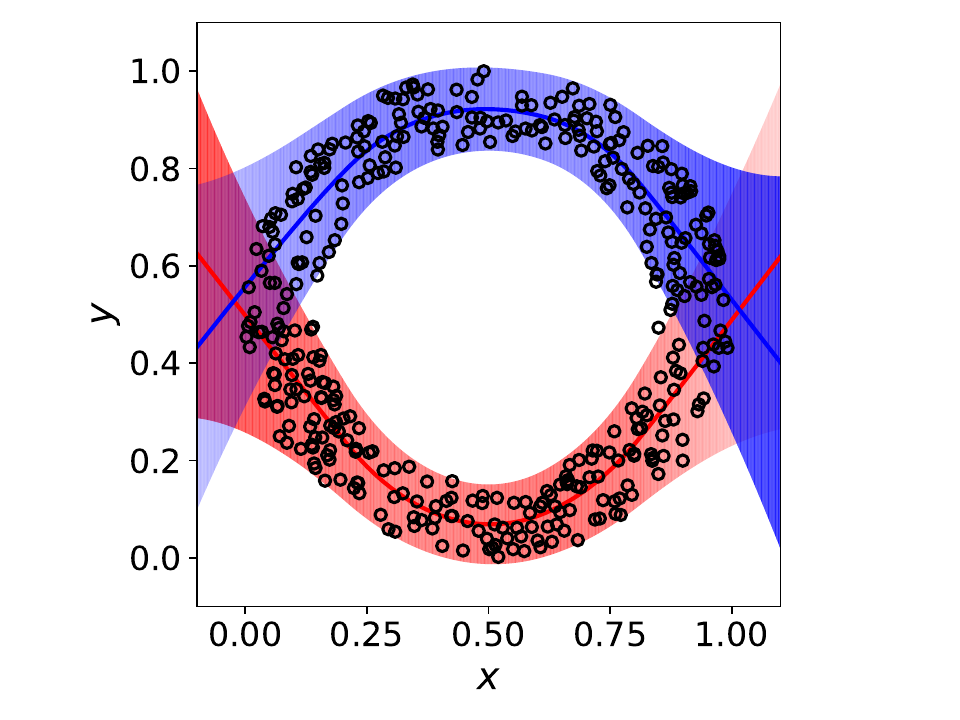}
  \end{subfigure}
  \begin{subfigure}{0.32\textwidth}
    \centering
    \includegraphics[width=\linewidth]{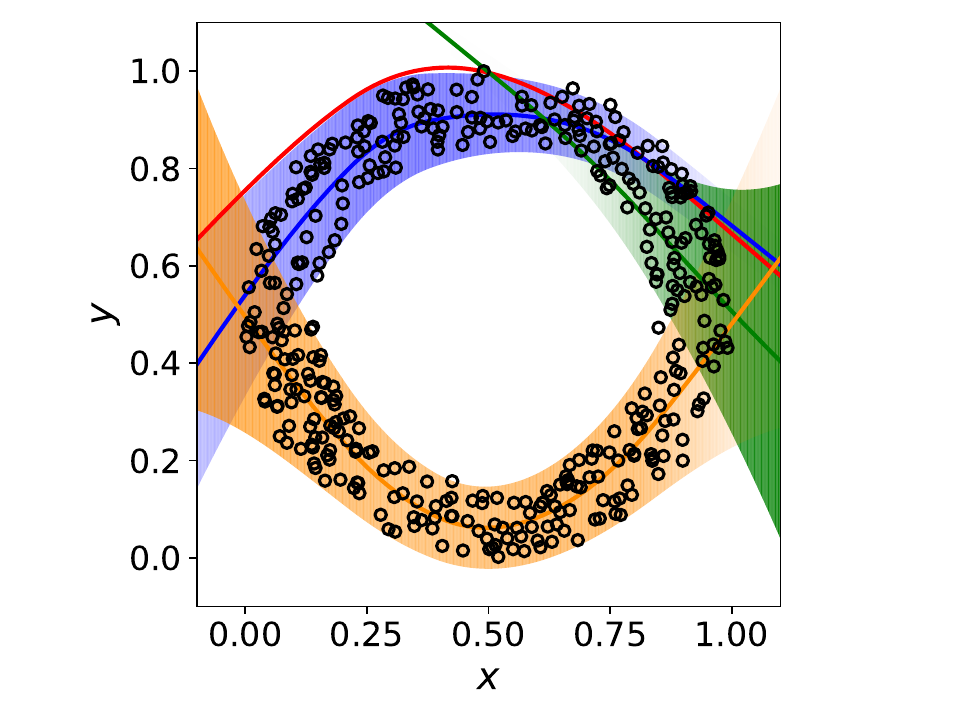}
  \end{subfigure}
  \begin{subfigure}{0.32\textwidth}
    \centering
    \includegraphics[width=\linewidth]{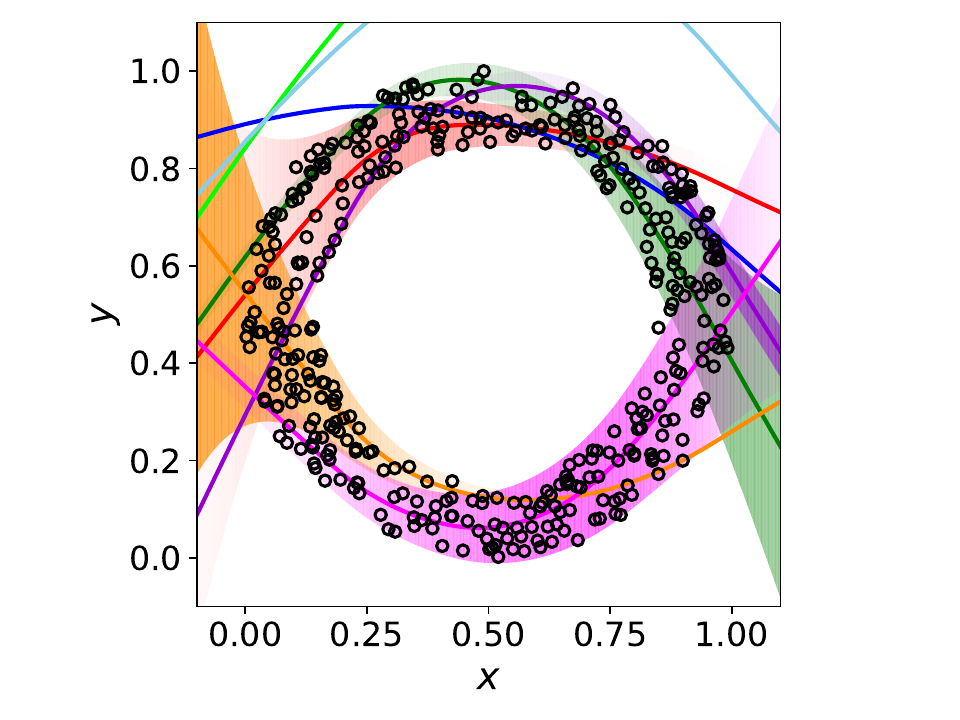}
  \end{subfigure}
  \begin{subfigure}{0.32\textwidth}
    \centering
    \includegraphics[width=\linewidth]{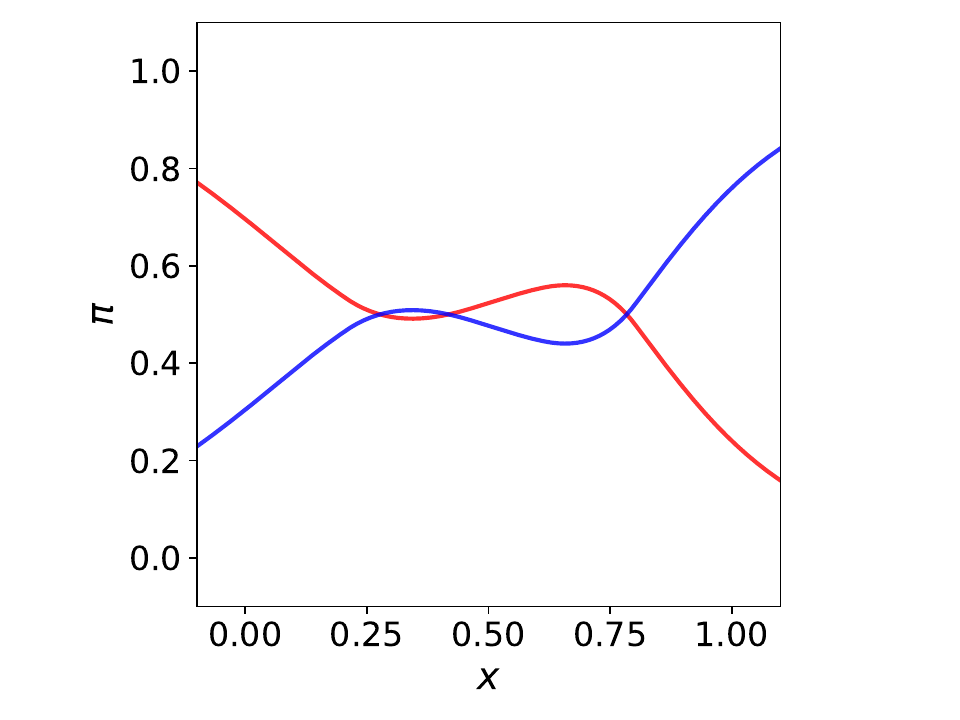}
  \end{subfigure}
  \begin{subfigure}{0.32\textwidth}
    \centering
    \includegraphics[width=\linewidth]{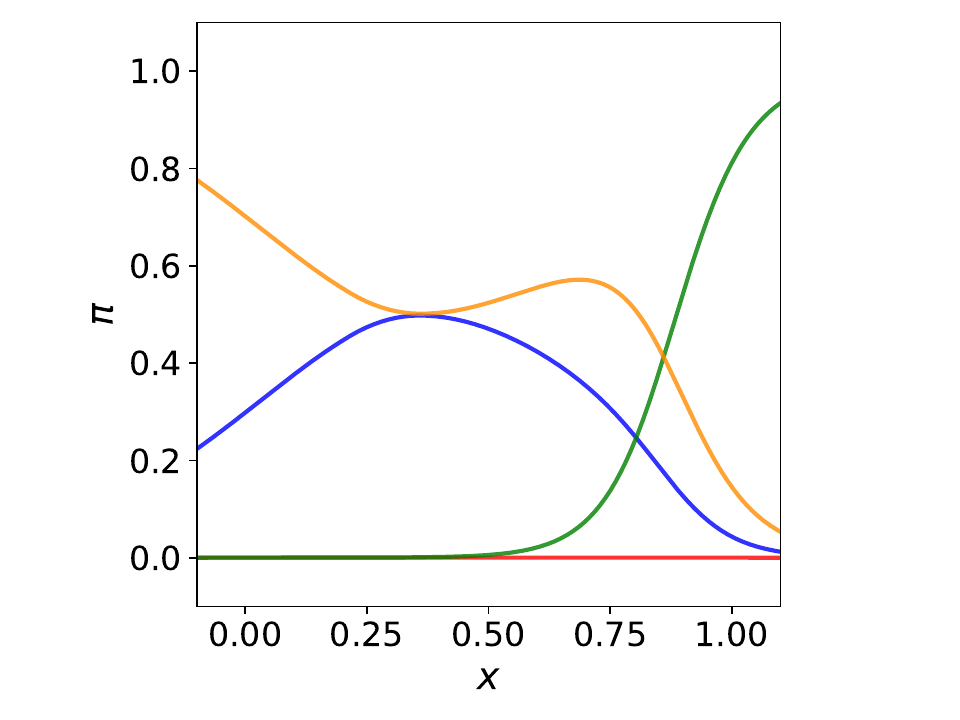}
  \end{subfigure}
  \begin{subfigure}{0.32\textwidth}
    \centering
    \includegraphics[width=\linewidth]{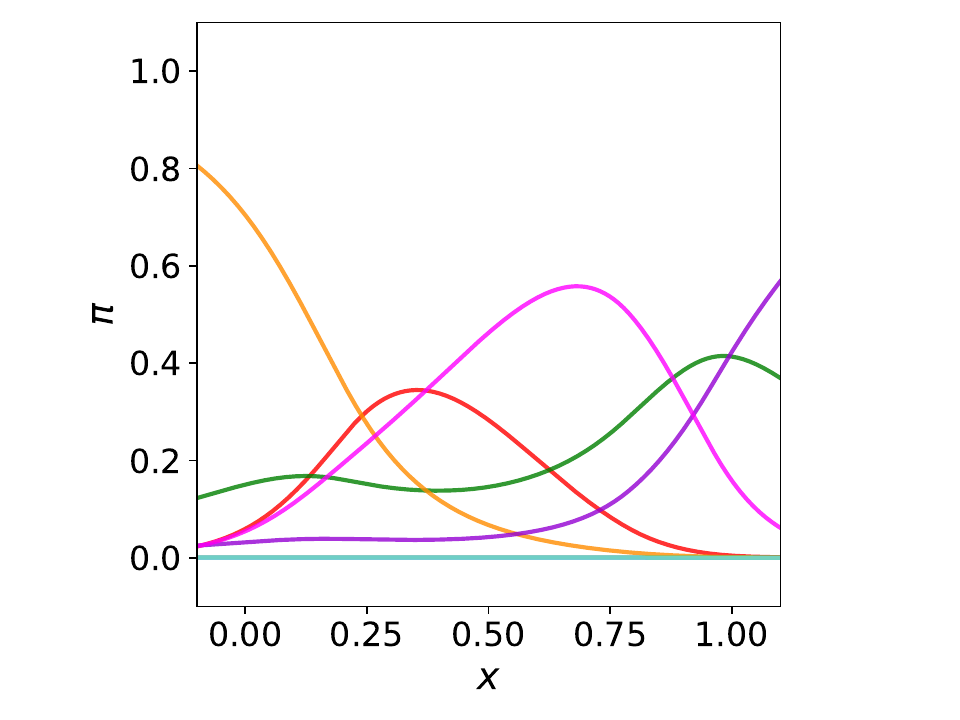}
  \end{subfigure}
  \begin{subfigure}{0.32\textwidth}
    \centering
    \includegraphics[width=\linewidth]{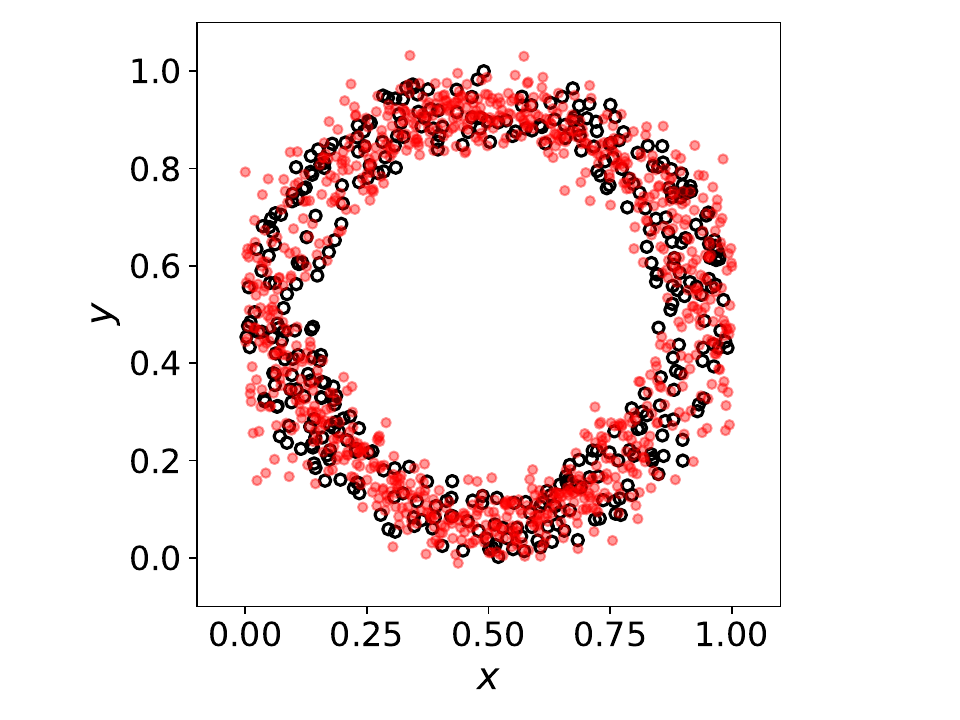}
  \end{subfigure}
  \begin{subfigure}{0.32\textwidth}
    \centering
    \includegraphics[width=\linewidth]{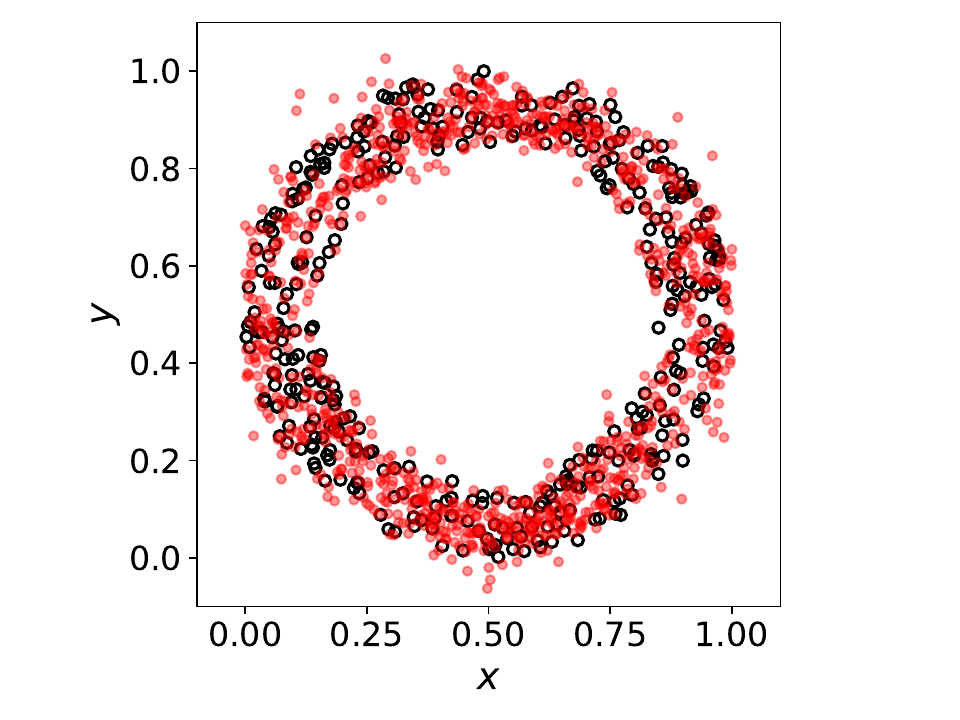}
  \end{subfigure}
  \begin{subfigure}{0.32\textwidth}
    \centering
    \includegraphics[width=\linewidth]{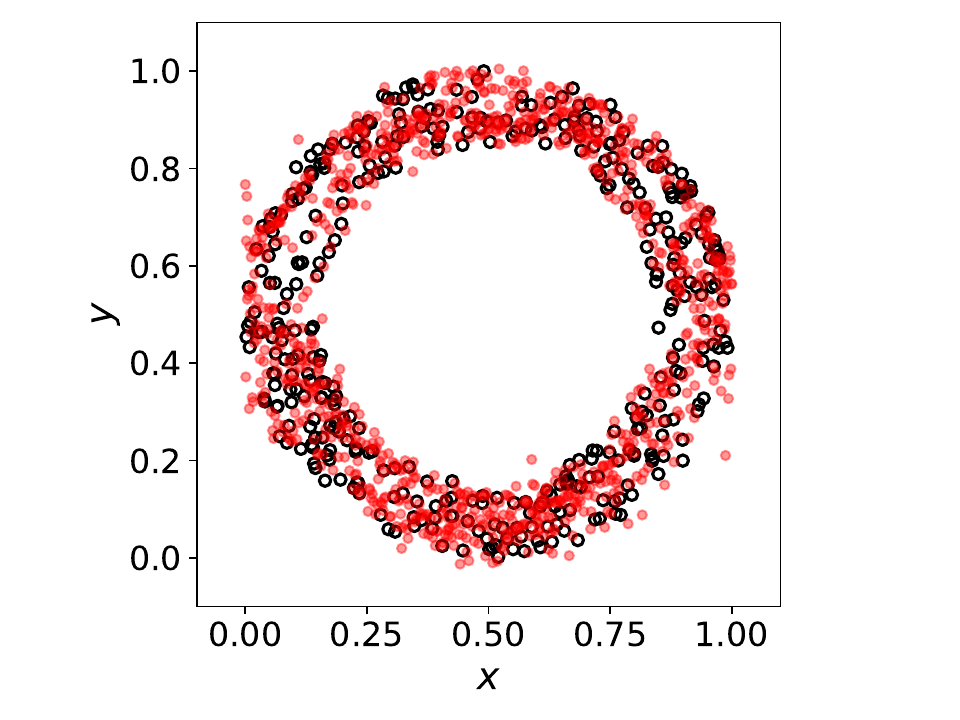}
  \end{subfigure}
  \hfill
    \caption{
    Results for the circular dataset with different numbers of mixture components after $5{,}000$ ADAM iterations.
    Top: learned conditional densities for $M=2,4,8$.
    Middle: corresponding mixture weights $\pi_m$.
    Bottom: $1{,}000$ samples drawn from each trained MDN model.
    All models are trained under identical settings.
    }
  \label{fig:circle_components}
\end{figure}

Figure~\ref{fig:circle_components} compares results obtained with $M=2$, $4$, and $8$ mixture components.
The top row shows that the learned conditional densities are visually similar across all configurations and recover the same multimodal structure of the data. The middle row reports the corresponding mixture weights $\pi_m$.
For larger $M$, several components are assigned negligible weights across the input domain: in the $M=4$ case, one component remains close to zero, while in the $M=8$ case, only five components attain nontrivial weights and the remaining three components are effectively suppressed. The bottom row shows $1{,}000$ samples drawn from each trained model, which overlap closely with the target circular distribution for all values of $M$.
Overall, these results show that comparable conditional representations are obtained across different choices of the number of mixture components.

\section{Conditional Flow Matching}

This appendix summarizes the CFM framework \cite{lipman2022flow} used to learn conditional generative models from data. CFM aims to approximate the conditional density $p_1(u|z,x)$ by constructing a time-dependent vector field that transforms a tractable base distribution into the target data distribution. Let $p_0(u)$ denote a simple reference distribution on $\mathbb{R}^d$ (e.g., a standard Gaussian), and let $p_1(u|z,x)$ denote the conditional distribution induced by the data. 

To link these distributions, CFM introduces a time-dependent, context-aware velocity field $\mathbf{v}_\theta(u,t,z,x): \mathbb{R}^d \times [0,1] \times \mathcal{Z} \times \mathcal{X} \to \mathbb{R}^d$, which defines a deterministic flow through the ordinary differential equation (ODE):
\begin{equation}
\label{eq:cfm_ode}
\frac{du(t)}{dt} = \mathbf{v}_\theta(u(t),t,z,x), \qquad u(0) \sim p_0.
\end{equation}
For a fixed context $(z,x)$, integrating Equation~\ref{eq:cfm_ode} from $t=0$ to $t=1$ defines a push-forward map that transforms the base distribution $p_0$ into the target approximation of $p_1(\cdot|z,x)$.

Training is performed using a simulation-free objective based on a prescribed interpolation between noise and data samples. Given a data sample $u_1 \sim p_1(\cdot|z,x)$ and an independent draw $u_0 \sim p_0$, we define a conditional probability path as a linear bridge:
\begin{equation}
\label{eq:cfm_bridge}
u_t = (1-t)u_0 + t u_1, \qquad t \in [0,1],
\end{equation}
which possesses a constant pathwise velocity $\mathbf{w}_t := \frac{du_t}{dt} = u_1 - u_0$. The parameters $\theta$ are optimized by matching the learned velocity field $\mathbf{v}_\theta$ to this target velocity $\mathbf{w}_t$. Specifically, by sampling $(z,x,u_1)$ from the dataset, $u_0 \sim p_0$, and $t \sim \mathrm{Uniform}(0,1)$, the model is trained by minimizing the expected $L_2$ regression loss:
\begin{equation}
\label{eq:cfm_loss}
\mathcal{L}(\theta) = \mathbb{E}_{t,z,x,u_0,u_1} \left[ \big\| \mathbf{v}_\theta(u_t,t,z,x) - (u_1 - u_0) \big\|^2 \right].
\end{equation}
Once trained, realizations of the conditional random field are generated by drawing $u(0) \sim p_0$ and solving Equation~\ref{eq:cfm_ode} forward in time using a numerical integrator, such as an Euler scheme, while holding the context $(z,x)$ fixed.

\section{Stationary Solutions of the Fokker–Planck Equation}\label{appendix:stationary_solution}
Consider the one-dimensional It\^{o} stochastic differential equation (SDE)
\begin{equation}
\label{eq:sde_u2_general}
du_2 = b(u_2;u_1)dt + a_3 dB(t),
\end{equation}
where $u_1$ is treated as a frozen parameter (e.g., a slow variable), $B(t)$ is a standard Brownian motion, and $a_3>0$ is a constant noise amplitude (additive noise).

Let $p(u,t|u_1)$ denote the probability density function (pdf) of $u_2(t)$. The corresponding Fokker--Planck equation is as follows \cite{karras2022elucidating}:
\begin{equation}
\label{eq:fpe_u2}
\partial_t p
= -\partial_u\!\big(b(u;u_1)\,p\big)
+ \frac{a_3^2}{2}\,\partial_{uu}p.
\end{equation}
At stationarity, we have:
\begin{equation}
\label{eq:fpe_stationary}
-\partial_u\!\big(b(u;u_1)\,p_\infty(u| u_1)\big)
+ \frac{a_3^2}{2}\,\partial_{uu}p_\infty(u| u_1) = 0,
\end{equation}
By introducing the probability flux $J$ as follows
\begin{equation}
\label{eq:flux_def}
J(u)
:= b(u;u_1)\,p_\infty(u| u_1) - \frac{a_3^2}{2}\,\partial_u p_\infty(u| u_1),
\end{equation}
The Equation~\ref{eq:fpe_stationary} can be written as a conservation law $\partial_u J(u)=0$, implying $J(u)\equiv J_0$ is constant. For a normalizable stationary density on $\mathbb{R}$ with $p_\infty(u| u_1)\to 0$ as $|u|\to\infty$, the steady flux must vanish, i.e., $J_0=0$. Therefore,
\begin{equation}
\label{eq:ode_stationary}
\partial_u p_\infty(u| u_1)
= \frac{2}{a_3^2}\,b(u;u_1)\,p_\infty(u| u_1),
\end{equation}
which yields
\begin{equation}
\label{eq:stationary_density_general}
p_\infty(u_2| u_1)
\propto
\exp\!\left(\frac{2}{a_3^2}\int_{s=-\infty}^{u_2} b(s;u_1)\,ds\right).
\end{equation}
The proportionality constant is determined by normalization $\int_{\mathbb{R}} p_\infty(u| u_1)\,du = 1$ whenever the integral is finite.

In the problem in Section~\ref{sec:sec:spde}, the drift function is $b(u_2;u_1)=-(-1 + 0.2\,u_1 + 4\,u_2(-1 + u_2^2))$. This expression is substituted into Equation~\ref{eq:stationary_density_general} and the corresponding
normalization factor is computed numerically using the trapezoidal rule:

\begin{equation}
    p_\infty(u_2|u_1)
\propto
\exp\!\left(
-\frac{2}{a_3^2}
\Big(
(u_2^2 - 1)^2 + (0.2u_1 - 1)u_2
\Big)
\right).
\end{equation}

\end{document}